%% file: main_camera_ready.tex
\documentclass{article}

  \usepackage[main, final]{neurips_2026}

\usepackage[utf8]{inputenc}
\usepackage[T1]{fontenc}
\usepackage{hyperref}
\usepackage{url}
\usepackage{booktabs}
\usepackage{amsfonts}
\usepackage{nicefrac}
\usepackage{microtype}
\usepackage{xcolor}

\usepackage{amsmath,amssymb}
\usepackage{graphicx}
\usepackage[export]{adjustbox}
\usepackage{tabularx}
\usepackage{multirow}
\usepackage{makecell}
\usepackage{enumitem}
\usepackage{xspace}
\usepackage[most]{tcolorbox}
\usepackage{float}
\usepackage{placeins}
\usepackage{cleveref}

\usepackage[disable]{todonotes}
\usepackage{xcolor}
\usepackage{etoc}
\usepackage{wrapfig} 
\usepackage{subcaption}

\addtocontents{toc}{\protect\setcounter{tocdepth}{-1}}

\usepackage{xcolor}

\usepackage{diagbox}

\newcommand{\method}{AgentArk\xspace}

\title{AgentArk: Distilling Multi-Agent Intelligence\\ into a Single LLM Agent}

\author{%
  Yinyi Luo$^{1}$\thanks{Contact author: \texttt{yinyil@andrew.cmu.edu}.}, Yiqiao Jin$^{3}$, Weichen Yu$^{1}$, Mengqi Zhang$^{2}$, Srijan Kumar$^{3}$, Xiaoxiao Li$^{5}$,\\
  \textbf{Weijie Xu$^{4}$}\thanks{Corresponding authors: \texttt{jdw@wm.edu} and \texttt{weijiexu@amazon.com}.}, \textbf{Xin Chen$^{4}$}, \textbf{Jindong Wang$^{2}$}\footnotemark[2]\\
  $^{1}$Carnegie Mellon University \quad
  $^{2}$William \& Mary \quad
  $^{3}$Georgia Institute of Technology\\
  $^{4}$Amazon \quad
  $^{5}$University of British Columbia\\
}

\begin{document}

\maketitle

\begin{abstract}

While large language model (LLM) multi-agent systems achieve superior reasoning performance through iterative debate, practical deployment is limited by their high computational cost and error propagation.
This paper proposes \textbf{\method}, a novel framework to distill multi-agent dynamics into the weights of a \textit{single} model, effectively transforming explicit test-time interactions into implicit model capabilities.
This equips a single agent with the intelligence of multi-agent systems while remaining computationally efficient.
Specifically, we investigate three hierarchical distillation strategies across various models, tasks, scaling, and scenarios: reasoning-enhanced fine-tuning; trajectory-based augmentation; and process-aware distillation.
By shifting the burden of computation from inference to training, the distilled models preserve the efficiency of one agent while exhibiting strong reasoning and self-correction performance of multiple agents.
They further demonstrate enhanced robustness and generalization across diverse reasoning tasks.
We hope this work can shed light on future research on efficient and robust multi-agent development.
Our code is available at \url{https://github.com/AIFrontierLab/AgentArk}.
\end{abstract}

\section{Introduction}


Multi-agent Systems (MAS), where multiple models interact through debate~\citep{du2023improving, hegazy2024diversity, eo2025debate}, critique~\citep{lan2024training, yu2025table}, and consensus~\citep{chen2024reconcile}, have demonstrated remarkable success in complex reasoning tasks~\citep{guo2024large}.
By structuring reasoning as a multi-turn and multi-role dialogue, MAS can explore diverse hypotheses, uncover logical errors~\citep{wang2024aegis}, and iteratively refine solutions~\citep{wan2025mamm}.
However, this collaborative power is a double-edged sword that introduces systemic risks including \emph{computational overhead} and \emph{vulnerability amplification}.
First, the reliance on multi-role and multi-turn dialogue causes inference latency and computational overhead to grow rapidly~\citep{wang2025mars, kim2025cost}.
In densely connected networks, computation can grow quadratically with the number of agents, making MAS prohibitively expensive for real-time \citep{kim2025towards, ignise2024applications}.
Second, while MAS can correct errors, they can also escalate them. In high-density interactions, individual biases or hallucinations can propagate and amplify across the group, leading to collective failures in robustness and safety~\citep{he2025comprehensive, nguyen2025social}.

These challenges naturally raise a fundamental question: \emph{Can a single model internalize the reasoning benefits of MAS without their high inference-time cost and collaborative vulnerabilities?}


The collective power of MAS, i.e., \emph{inference-time compute}, suggests that the gains can possibly be ``shifted forward''---internalized by a single model during offline learning~\citep{wang2025mas, chen2025optima, liu2025adaptive}.
While prior work shows that single models can absorb certain MAS reasoning benefits~\citep{li2026single, li2025chain, chen2024magdi, zhou2025debate}, they are often limited to imitating final answers~\citep{han2024llm} or exhibiting  shallow interaction traces~\citep{li2023camel, lin2025creativity}, failing to reproduce the core \emph{iterative conflict-and-refinement dynamics} that underlie MAS reasoning \citep{li2024survey}.
Notably, recent evidence suggests that the structural design of MAS may play a secondary role in observed gains.
\citep{kim2025towards} demonstrate that removing or perturbing explicit agent structures leads to only marginal performance degradation, revealing the essential contribution of MAS lies in the \emph{reasoning dynamics} they induce, rather than in the interaction schema itself.
Similarly, \citep{ke2026mas} systematically studies orchestration strategies under controlled benchmarks, showing that performance is largely driven by induced reasoning behaviors rather than agent topologies or protocols.

\begin{figure*}[t!]
    \centering
    \includegraphics[width=.9\linewidth]{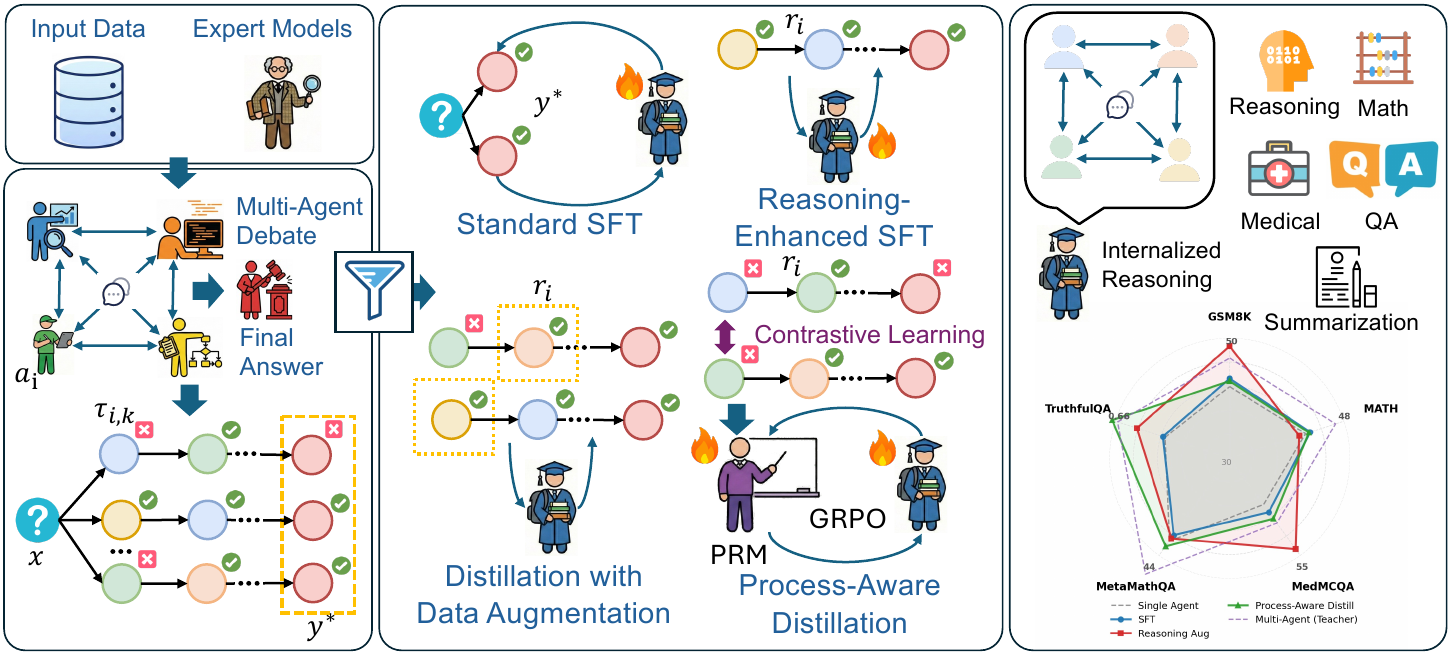}
    \caption{\textbf{Overview of \method}: (1) \textbf{Data Generation Through Multi-Agent Debate} to produce diverse reasoning trajectories; (2) \textbf{Knowledge Extraction} to filters for high-quality corrective traces; and (3) \textbf{Distillation} utilizing Standard SFT, Reasoning-enhanced SFT, Distillation with Data Augmentation, and Process-Aware Distillation (PRM + GRPO).
    }
    \label{fig:pipeline}
    \vspace{-.2in}
\end{figure*}

Our pilot study showed that fine-tuning only on final answers can lead to overfitting on task-specific mapping with limited generalization (Appendix \ref{app:sft_results}).
Therefore, a single agent should learn to internalize the reasoning dynamics, allowing for generating, evaluating, and refinement with one forward pass.
This perspective also resonates with human cognition: individuals can internalize group reasoning strategies and reproduce collective wisdom independently \citep{toyokawa2019social, navajas2018aggregated,curcseu2015cognitive}.
In this paper, we present a comprehensive investigation of multi-agent distillation for reasoning tasks.
Particularly, we propose \textbf{\method} (\Cref{fig:pipeline}), a general distillation framework that transfers MAS reasoning dynamics into a single model, without relying on handcrafted interaction or task-specific supervision.\footnote{We only focus on reasoning tasks; tool use~\citep{qin2024tool} and memory management~\citep{zhang2025survey} are for future work. We mainly consider the popular homogeneous setting where all agents share the same LLM backbone~\citep{eo2025debate, chen2023multi}. Heterogeneous results are in \S\ref{sec-exp-multi}.}

\method integrates three progressively deeper distillation strategies:
1) Reasoning-Enhanced Outcome–Based Supervision \emph{(R-SFT)}: Training the model on the final consensus reached by the agent group while leveraging reasoning trajectories;
2) Reasoning Trajectory-based Data Augmentation \emph{(DA)}: Distilling the diverse reasoning chains derived from collective interaction, allowing the model to learn various logical strategies and ways of thinking;
and 3) Process-Aware Distillation \emph{(PAD)}: Using process reward model to train the agent to internalize the critique-and-revision dynamics, enabling a single agent to emulate the dialectical reasoning of multi-agent debates within a single forward pass.
We conduct extensive experiments by varying LLM backbones, teacher-student roles, datasets and various tasks, scaling, and evaluation settings.
The following are our key findings:
\vspace{-.1in}
\begin{enumerate}[leftmargin=1em]
\setlength\itemsep{-.1em}
    \item \textbf{\method enables a single agent to acquire multi-agent reasoning ability.} Our extensive experiments show that all three reasoning-centric distillation methods can boost the performance of single agents. Combination of approaches can achieve further improvement. (\S\ref{sec-exp-result})

    \item \textbf{PRM capacity matters more than student model size, while student capacity bounds multi-agent gains.} High-capacity PRMs enable strong improvements even for small students, whereas weak PRMs limit gains. Scaling teacher agents mainly benefits larger students, with diminishing returns for smaller ones. (\S\ref{sec-exp-result}, \S\ref{sec-exp-scaling})

    \item \textbf{Reasoning quality outweighs quantity.} Simply adding more reasoning trajectories does not improve performance, while PAD's high-signal process supervision yields stable gains. (\S\ref{sec-exp-scaling})

    \item \textbf{Process-aware distillation improves reasoning behavior, not just accuracy.} PAD models exhibit better step decomposition, self-checking, and error correction than RSFT and RA. (\S\ref{sec-exp-case})

    \item \textbf{\method improves generalization and robustness.} Distilled models transfer reliably to unseen and robustness benchmarks. (\S\ref{sec-exp-robust})

    \item \textbf{\method extends to multimodal and heterogeneous LLMs and other MAS algorithms.} (\S\ref{sec-exp-multi})
    \vspace{-.1in}
\end{enumerate}

\vspace{-.1in}
\textbf{Contributions.}
(1) To our best knowledge, \method is the \emph{first} comprehensive framework to explore various strategies for MAS distillation.
(2) We construct a scalable distillation data generation pipeline and framework agnostic to MAS strategies, which will be released to facilitate future research on reasoning distillation.
(3) We perform extensive evaluation of MAS distillation from various perspectives, providing insights for future research.

\vspace{-2mm}
\section{Related Work}
\label{sec:related}

\vspace{-1mm}
\textbf{Multi-Agent Systems.}
MASs have emerged as an effective paradigm for enhancing LLM reasoning by enabling multiple agents to interact \citep{pmlr-v267-yuan25l, chen2024reconcile, eo2025debate, du2023improving, wei2026agentic}. By organizing reasoning as explicit multi-turn interactions, they can explore diverse solution paths, detect errors, and iteratively refine predictions, leading to strong performance on complex reasoning tasks \citep{wang2024aegis,wang2025companioncast,wang2026mascot}. However, MAS rely on inference-time coordination among multiple agents, which incurs substantial computational cost and latency \citep{shi2026learning, ma2026maestro, wang2025survey,wang2025anymac}.
In addition, agent roles, interaction protocols, and evaluation criteria are typically designed for specific tasks, limiting the applicability in resource-constrained or real-time settings.
Beyond efficiency, recent work \citep{kim2025towards} examined the structural sensitivity of MAS, showing that performance is often robust to substantial simplifications or perturbations of agent structures.

\textbf{Distillation of Multi-Agent Reasoning.}
To avoid the above issues, recent work has explored distilling multi-agent reasoning into a single model \citep{liu2025language, zhao2024we, wang2025fedlma, pan2025knowledge}. Early approaches supervise student models using the final outputs of agent groups or simplified interaction traces \citep{liu2025structured, kang2025distilling}. More recent methods transfer richer supervision signals, including graph-based interaction modeling \citep{chen2024magdi}, skill selection frameworks \citep{li2026single}, debate-derived preference supervision \citep{zhou2025debate} and end-to-end agentic reinforcement learning \citep{li2025chain}. Despite their effectiveness, they often depend on task-specific agent designs, predefined interaction structures, or environment-dependent reward functions, restricting generalization across tasks and limit scalability~\citep{wan2025beyond}.
In contrast, \method abstracts away agent roles and interaction structures by distilling interaction reasoning processes at the process level, enabling a single student model to generalize across tasks and agent configurations without task-specific design.

\vspace{-.1in}
\section{Method}
\label{sec:method}



As shown in \Cref{fig:pipeline}, \method consists of three phases: (1) \textbf{Data Generation via Multi-Agent Interaction}, where teacher models generate diverse reasoning trajectories. Here, we leverage the iterative reflection and error-correction trajectories inherent in LLM debates~\citep{du2023improving}. (2) \textbf{Knowledge Extraction}, where successful and corrective traces are generated and filtered; and (3) \textbf{Hierarchical Distillation}, where the student is trained via supervised learning and process-level RL.
We explore three paradigms: (1) Reasoning-enhanced Supervised Fine-Tuning (RSFT), (2) Data Augmentation via Diverse Extraction (DA), and (3) Process-Aware Distillation using PRM and GRPO.

\subsection{Data Generation and Knowledge Extraction}

We adopt debate~\citep{liang2024encouraging} as the MAS mechanism for data generation with rich reasoning dynamics.
Interactions where individuals reflect, revise, and converge have long been recognized as hallmarks of intelligent behavior, and debate naturally surfaces processes such as self-correction, error discovery, and cross-examination.
Prior work on debate~\cite{du2023improving} demonstrates that structured disagreement can enhance both factual accuracy and reasoning quality.
Results on other MAS algorithms are in \S\ref{sec-exp-multi}.

\textbf{Debate Setup.}
For each input $x$, we initialize $n$ debating agents $\mathcal{A} = \{a_1, a_2, \dots, a_n\}$ sharing the same LLM. The agents engage in a $K$-round interaction. In each round, each agent $a_i$ generates a reasoning trace $\tau_{i, k}$ based on the problem $x$ and the previous traces of its peers $\{\tau_{j, k-1}\}_{j \neq i}$.
This setup encourages agents to identify logical inconsistencies in others' arguments and iteratively refine their own reasoning traces. This process continues until $K$ rounds is reached or a consensus emerges. The result is a comprehensive debate log $\mathcal{L}_x$ containing a set of final reasoning traces $\{r_1, r_2, \dots, r_n\}$ and their corresponding answers.

\textbf{Correctness-First Trajectory Selection.}
While traditional distillation typically utilizes error-free paths,
we prioritize \textbf{corrective trajectories}, reasoning traces where an agent initially proposed an incorrect step but recognize successfully pivoted to a correct answer $y^*$ after receiving critiques. For each task, we extract 1) the final consensus answer $y^*$, verified against ground-truth labels; 2) the intermediate reasoning traces $\{r_i\}$ that successfully lead to $y^*$.

\textbf{Knowledge Extraction.}
We extract multiple answer-consistent yet structurally diverse reasoning trajectories from debates, capturing both correct solutions and explicit self-correction behaviors.

\subsection{Distillation Methods}
\label{sec-method-method}

We explore three strategies to supervise the student model $\pi_\theta$, aiming to distill not only correct solutions but also concise and structured reasoning~\citep{zhong2025rethinking}.

\subsubsection{Reasoning-Enhanced SFT}
Reasoning-enhanced SFT not only takes the final answers, but also the reasoning traces as supervision.
The student model is trained to maximize the likelihood of the raw multi-agent reasoning traces. Given an input $x$ and a successful trace $r = (r_1, \dots, r_{n})$ to final answer $y^*$, the objective is:

\begin{equation}
\label{eq-sft}
\mathcal{L}_{\text{SFT}}(\theta) = - \mathbb{E}_{(x, r, y^*) \sim \mathcal{D}} \mathcal{L}_{\text{res}} + \mathcal{L}_{\text{ans}},
\end{equation}
where
\begin{align}
    \mathcal{L}_{\text{res}} &= \sum_{t=1}^{|r|} \log p_\theta(r_t \mid r_{<t}, x) \tag{Reasoning}\label{eq:l_res}, \\
    \mathcal{L}_{\text{ans}} &= \log p_\theta(y^* \mid r, x). \tag{Answer}\label{eq:l_ans}
\end{align}

The objective consists of two components: a reasoning loss $\mathcal{L}_{\text{res}}$, which optimizes the model's ability to generate a coherent sequence of intermediate rationales $r$, and an answer loss $\mathcal{L}_{\text{ans}}$, which ensures the final prediction $y^*$ is grounded in both the input context $x$ and the preceding reasoning path.
SFT assesses the student's capacity for basic imitation of the multi-agent reasoning style, testing whether it can maintain structural consistency across the entire generation sequence.

\subsubsection{Distillation with Data Augmentation}
\label{sec-method-da}

To fully exploit the diversity and multi-perspective nature of the multi-agent debate, we implement a \textit{Correctness-First Diverse Extraction} strategy.
For each problem $x$, we filter the agents to define a set of successful contributors $\mathcal{A}_{\text{correct}} \subseteq \mathcal{A}$.
We then utilize a high-capacity teacher LLM as a ``distiller'' to parse the raw debate logs of these agents. The distiller is tasked with extracting $k \in \{1, 2, 3\}$ reasoning trajectories that are:
(1) \textbf{Correct:} They must lead strictly to the ground-truth $y^*$.
(2) \textbf{Diverse:} The teacher is prompted to select traces that employ distinct mathematical identities, different logical heuristics, or varied starting assumptions (details in Appendix~\ref{app:data-generation}).
The student is trained on this augmented set $\mathcal{D}_{\text{aug}}$ using:
\begin{equation}
\mathcal{L}_{\text{Aug}}(\theta) = - \frac{1}{k} \sum_{i=1}^{k} \sum_{t=1}^{T} \log p_\theta(y_{t} \mid y_{<t}, r_i, x).
\end{equation}
This forces model to learn multiple valid paths to the same solution for robustness and generalization.

\subsubsection{Process-Aware Distillation}
The third method treats distillation as a reinforcement learning problem, using a Process Reward Model (PRM)~\citep{lightman2023let, wang2024math} for granular, step-level supervision that reinforces the intermediate logical transitions found in multi-agent debates.
Specifically, we train a PRM $R_\phi$ to predict the probability that a reasoning step is correct.
To ensure the PRM reliably captures the nuances of the debate, we use a two-stage curriculum:
\textbf{Stage I: Feature Alignment (Backbone Frozen).} We initialize $R_\phi$ with the weights of the student model and freeze all layers except for the final one and the reward head. This prevents the loss of pre-trained linguistic features while the reward head learns to map existing representations to the correctness labels $z_t \in \{0, 1\}$.
\textbf{Stage II: Full Specialization (Backbone Unfrozen).} We unfreeze the entire backbone for end-to-end fine-tuning. This allows the model to develop specialized internal attention patterns for detecting logical fallacies.


We design the Process Reward Model (PRM) to provide step-level supervision via a \textbf{contrastive reward objective} rather than standard binary cross-entropy.
This encourages the model to assign higher rewards to reasoning steps that are more consistent with the multi-agent debate consensus, reflecting relative correctness rather than absolute labels (details in Appendix~\ref{app:contrastive_loss}).

\textbf{Reinforcement Learning via GRPO.}
Finally, we fine-tune the student policy $\pi_\theta$ using Group Relative Policy Optimization (GRPO)~\citep{shao2024deepseekmath}. Given an input $x \sim \mathcal{D}$, we sample a group of $G$ reasoning outputs ${o_1, \dots, o_G}$ from a fixed behavior policy $\pi_{\text{old}}$, which is a snapshot of the student policy before the current update.
GRPO optimizes the policy by comparing the rewards of these outputs within the group, removing the need for a separate value function:
\begin{equation}
\mathcal{J}(\theta) = \mathbb{E}_{x \sim \mathcal{D}, \{o_i\} \sim \pi_{\text{old}}} \left[ \frac{1}{G} \sum_{i=1}^G \mathcal{L}_{i}(\theta) - \beta \mathbb{D}_{\text{KL}}(\pi_\theta \| \pi_{\text{ref}}) \right]
\end{equation}
where $\pi_{\text{ref}}$ denotes a fixed reference policy used to regularize the update, $\beta$ is the KL penalty coefficient, and $\mathcal{L}_{i}(\theta)$ is the clipped surrogate objective for output $o_i$, the surrogate objective for each output $o_i$ is:
\begin{equation}
\mathcal{L}_{i}(\theta) = \min \left( \rho_i(\theta) \hat{A}_i, \text{clip}(\rho_i(\theta), 1-\epsilon, 1+\epsilon) \hat{A}_i \right).
\end{equation}

In this formulation, $\rho_i(\theta) = \frac{\pi_\theta(o_i|x)}{\pi_{\text{old}}(o_i|x)}$ denotes the probability ratio. The advantage $\hat{A}_i$ ensures \textbf{reward consistency} by normalizing the step-wise aggregate score $R_{\phi}(o_i)$ from the trained PRM:
$
\hat{A}_i = (R_{\phi}(o_i) - \mu_R)/\sigma_R,
$
where $\mu_R$ and $\sigma_R$ are the mean and standard deviation of the PRM scores in group $G$ sampled outputs.

\section{Experiments}
\label{sec-exp}

\subsection{Experimental Setup}
\label{sec-exp-setup}

\textbf{Models.}
We conduct evaluation across three major model families: (1) \textbf{Qwen 3}~\cite{yang2025qwen3}: Qwen3-32B, Qwen3-8B, Qwen3-1.7B, and Qwen3-0.6B;
(2) \textbf{Gemma 3}~\cite{team2025gemma}: Gemma3-27B-it and Gemma-7B; and (3) \textbf{Llama 3}~\cite{dubey2024llama}: Llama3-8B-Instruct.
Particularly, we perform distillation from larger models (Qwen3-32B, and Gemma3-27B-it) to smaller ones (Qwen3-8B, Qwen3-1.7B, Qwen3-0.6B, Llama3-8B, and Gemma-7B) for comprehensive evaluation.
We additionally evaluate multimodal LLM, heterogeneous agents, and other MAS algorithms as a preliminary study in \S\ref{sec-exp-multi}.

\textbf{Datasets.}
We adopt diverse benchmarks: (1) MATH~\citep{lightman2023let} and GSM8K~\citep{cobbe2021training} for \textit{mathematical reasoning};
(2) MedMCQA~\citep{pal2022medmcqa} for \textit{domain-specific knowledge};
(3) MetaMathQA (MMQA)~\citep{yu2023metamath} for augmented math;
and (4) QASPER~\citep{dasigi2021dataset}, HotpotQA~\citep{yang2018hotpotqa} and QMSum~\citep{zhong2021qmsum} for \textit{multi-hop and long-form reasoning}.

\textbf{Comparison Methods.}
We mainly compare three distillation methods in \S\ref{sec:method}: RSFT, DA, and PAD, with single-agent and the vanilla multi-agent debate as baselines.
Our primary study are based on $5$ agents following existing work~\citep{liang2024encouraging} and \S\ref{sec-exp-scaling} presents the scaling exploration.
By varying teacher-student models and training-test datasets on different distillation methods, we conducted $120$ experiments in total.
Accuracy on the test dataset is adopted as the primary evaluation metric.

\subsection{Main Results}
\label{sec-exp-result}

\begin{figure}[t!]
    \centering
    \begin{subfigure}{0.48\textwidth}
        \centering
        \includegraphics[width=\linewidth]{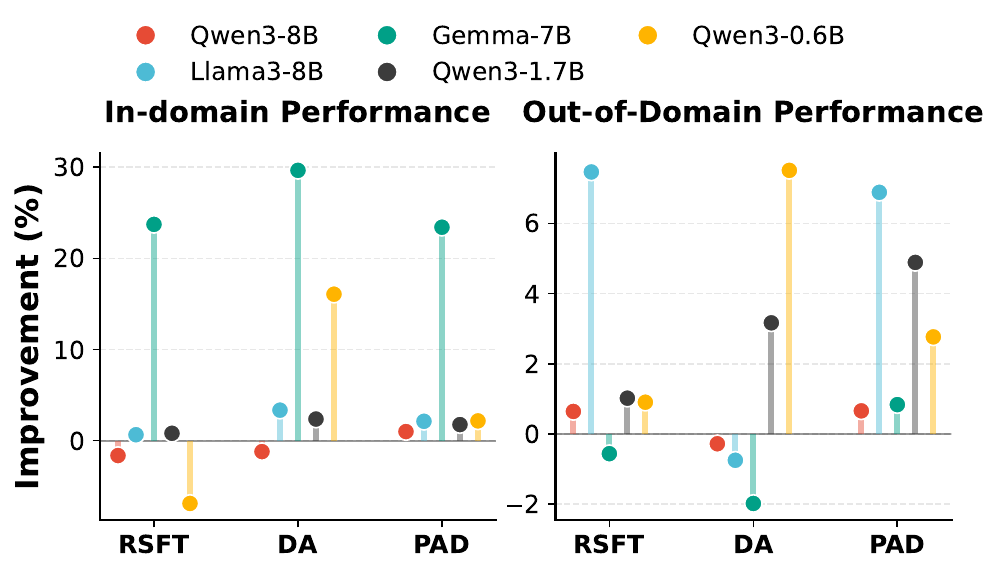}
        \caption{In-domain (left) vs OOD (right)}
        \label{fig:effectiveness_id_ood}
    \end{subfigure}
    \hfill
    \begin{subfigure}{0.48\textwidth}
        \centering
        \includegraphics[width=\linewidth]{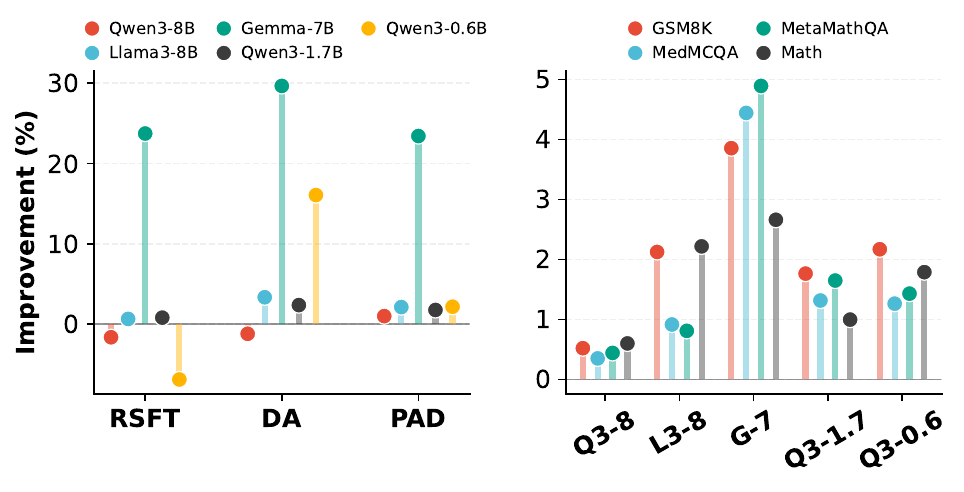}
        \caption{By datasets (left) and models (right)}
        \label{fig:target_model_dataset}
    \end{subfigure}
    \vspace{-.1in}
    \caption{Distillation from Qwen3-32B to different student models.}
    \label{fig-main-exp}
    \vspace{-.2in}
\end{figure}

\Cref{fig-main-exp} presents the results by distilling from Qwen3-32B to different student models across various datasets and the right-hand-side of \Cref{fig:pipeline} shows the average comparison results. Other results are in Appendix~\ref{app:all_models_results}.
Overall speaking, \method significantly improves the performance of a single agent by 4.8\% and only slightly worse than the vanilla MAS.
More insightful findings are as follows.






\textbf{Performance across distillation methods.}
\Cref{fig-main-exp} show the results tested on GSM8K and MedMCQA, respectively.
(1) \textbf{ID vs. OOD.} While distillation shows improvement in both settings, the gain in ID is more significant than in the OOD setting (30\% vs. 7\% maximum improvement and 4-6\% vs. 1-3\% on average).
This is expected as OOD is more challenging, suggesting more future work.
(2) \textbf{Methods.} Different methods show varied performance across datasets and tasks.
While RSFT and DA sometimes provide improvements, their gains are inconsistent and can fluctuate by dataset.
In contrast, PAD consistently yields performance improvements, demonstrating its robustness and reliable transfer of reasoning capabilities.
(3) \textbf{Compatibility.} Different distillation strategies are mutually compatible and can be composed to yield consistent gains (Appendix~\ref{app:combine_methods}).


\textbf{Performance across student models.}
We evaluate the generalization of all three distillation strategies by fixing the teacher model and varying the student model family.
The results are shown in \Cref{fig:effectiveness_id_ood} and \ref{fig:target_model_dataset} (left).
(1) \textbf{Same-family distillation (Qwen-3)}. When both teacher and student are drawn from the Qwen-3 family, distillation yields stable but relatively moderate improvements, with smaller variants (1.7B and 0.6B) consistently benefiting more than the 8B model, indicating that same-family distillation mainly alleviates capacity constraints rather than inducing substantial representational changes.
(2) \textbf{Cross-family distillation.} When distillation is performed across different model families, we observe larger and more consistent performance gains, particularly for Gemma-7B and LLaMA-3-8B, indicating that heterogeneous architectures benefit more from transferred reasoning signals.
(3) \textbf{Effect of PRM supervision.} Across both large and small student models, PRM-based distillation consistently improves performance on ID and OOD tasks, demonstrating that the learned gains are not limited to scale or data distribution.
Notably, the improvements persist under distribution shift, suggesting that PRM supervision primarily transfers reasoning behaviors rather than merely improving surface-level alignment to training data (Detailed results in Appendix~\ref{app:all_models_results}). We further conduct detailed PRM ablation studies to analyze its effects (Appendix~\ref{app:sec-append-ablation}).


\textbf{Performance across datasets.}
For a fixed student model, distillation consistently improves performance across all benchmark datasets, but the magnitude of gains varies (\Cref{fig-main-exp}).
MetaMathQA exhibits the largest improvements, followed by GSM8K, while Math shows moderate gains and MedMCQA benefits the least.
This pattern suggests that the observed improvements are primarily driven by enhanced reasoning capabilities rather than dataset-specific overfitting.
In particular, the strong gains on MetaMathQA and GSM8K indicate that these datasets contain reasoning-intensive problems, such as multi-step logical or arithmetic chains, which benefit most from transferred reasoning knowledge.
By contrast, Math, with more formulaic or domain-specific tasks, shows moderate improvement, and MedMCQA, which relies heavily on specialized medical factual knowledge, benefits the least, implying that distillation contributes less when reasoning is minimal.
Overall, this analysis highlights that our distillation strategies effectively enhance generalized reasoning skills, with larger impact on datasets that require complex reasoning rather than memorization.

\subsection{Scaling and Data Dynamics}
\label{sec-exp-scaling}

\begin{figure}[t!]
    \centering
    \begin{subfigure}{0.48\columnwidth}
        \centering
        \includegraphics[width=\linewidth]{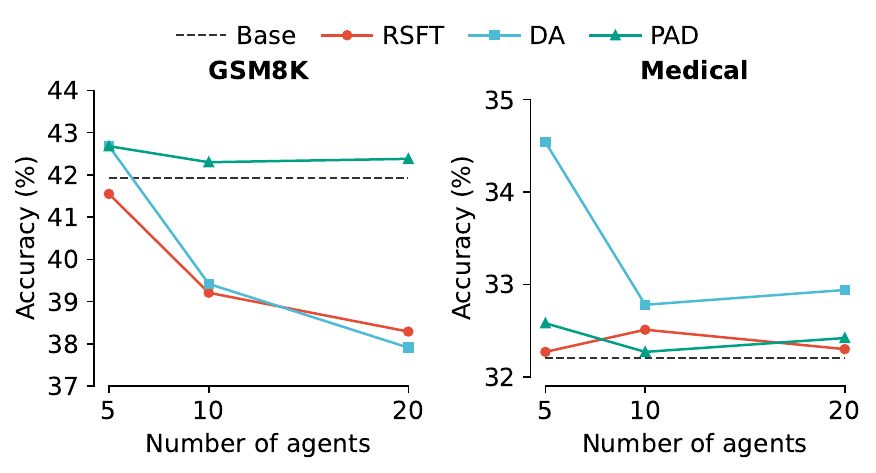}
        \vspace{-.1in}
        \caption{0.6B on GSM8K}
        \label{fig:agent_scale_06b}
    \end{subfigure}
    \hfill
    \begin{subfigure}{0.48\columnwidth}
        \centering
        \includegraphics[width=\linewidth]{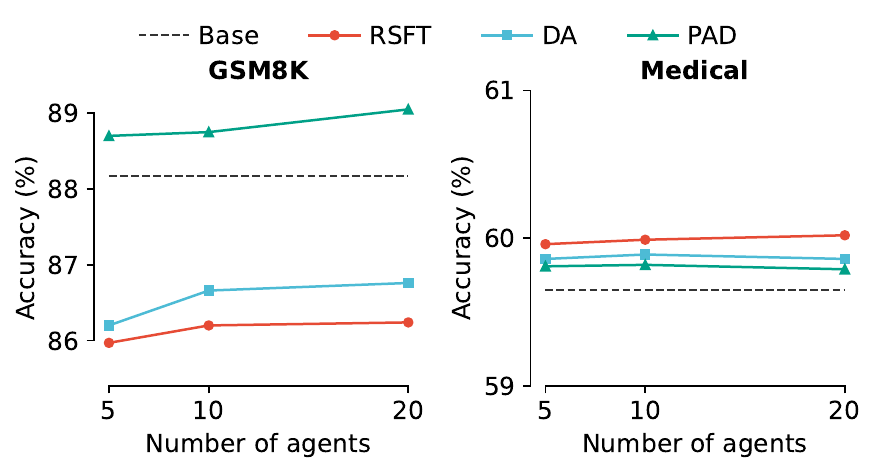}
        \vspace{-.1in}
        \caption{8B on GSM8K}
        \label{fig:agent_scale_8b}
    \end{subfigure}
    \vspace{-.1in}
    \caption{Agent scale ($5,10,20$) on distillation performance evaluated on GSM8K and MedMCQA.}
    \label{fig:agent_scale_exp}
    \vspace{-.2in}
\end{figure}

\textbf{Scaling the Number of Agents.}
While the primary experiments employ $5$ agents following existing work, we further scale to $10$ and $20$ agents to study how increasing teacher diversity and interaction complexity affects distillation.
We perform evaluation by distilling from a fixed Qwen3-32B model into two different student models: Qwen3-8B and Qwen3-0.6B.
The results are shown in \Cref{fig:agent_scale_exp}.

For the smaller student 0.6B, scaling $>5$ does not yield additional benefits and, in some cases, leads to performance degradation.
We attribute this to student's limited representational capacity: as the teacher ensemble becomes more diverse and produces more complex or longer reasoning trajectories, the student is unable to faithfully absorb and generalize this information.
By contrast, the larger Qwen3-8B benefits modestly from scaling. Performance improves consistently as the number of agents increases.
However, the incremental gains diminish at higher scales, indicating that once the student can effectively utilize the ensemble’s reasoning, additional agents offer limited benefit.
These results highlight that MAS distillation is bounded by student capacity: smaller models saturate quickly and may even struggle with overly diverse teacher signals, whereas larger models can better exploit richer supervision from multiple agents.
Overall, scaling teachers is effective only when matched to the student’s representational capacity (More results in Appendix~\ref{app:data-scaling}).




\begin{wrapfigure}{r}{0.5\textwidth}
    \centering
    \begin{subfigure}{0.48\linewidth}
        \centering
        \includegraphics[width=\linewidth]{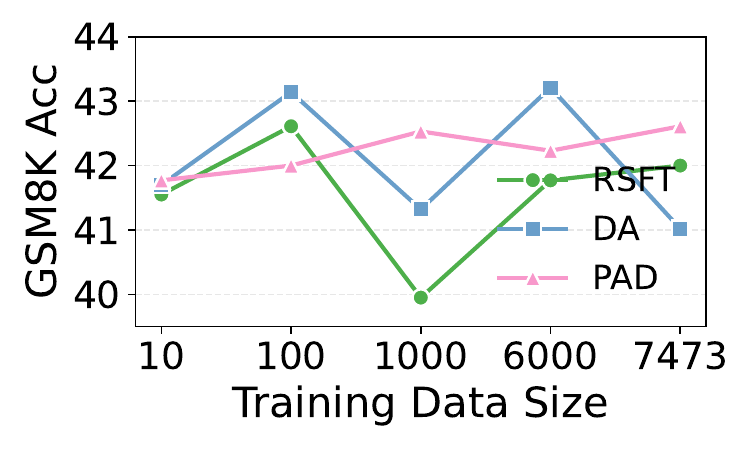}
        \caption{GSM8K}
        \label{fig:data_scale_gsm8k}
    \end{subfigure}
    \hfill
    \begin{subfigure}{0.48\linewidth}
        \centering
        \includegraphics[width=\linewidth]{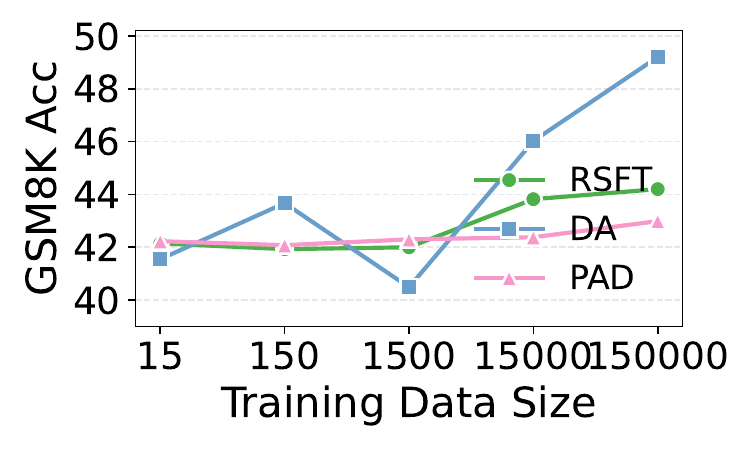}
        \caption{MetaMathQA}
        \label{fig:data_scale_meta}
    \end{subfigure}
    \vspace{-.1in}
    \caption{Data scaling results (distillation from Qwen3-32B to Qwen3-0.6B).}
    \label{fig:data_scale}
    \vspace{-.2in}
\end{wrapfigure}
\textbf{Data Quantity vs. Quality.}
The amount of training data is varied in \Cref{fig:data_scale}, showing that increasing training data does not lead to monotonic improvements.
For both RSFT and DA, performance exhibits high variance as data scale grows: moderate data sizes can yield gains, while further scaling often leads to stagnation or even degradation, particularly for GSM8K and MedMCQA.
In contrast, PAD demonstrates significantly more stable behavior across data scales.
They consistently achieve competitive performance and avoid the sharp fluctuations observed with raw data scaling.
This suggests that for capacity-limited students, reasoning quality rather than data volume is the primary bottleneck.
Excessive or noisy supervision from large-scale MAS outputs can overwhelm the target, whereas PRM-guided distillation preserves high-signal reasoning trajectories for more reliable transfer.

\textbf{Training Time.}
The offline training cost is shown in Appendix~\ref{app:computation-cost}, which is mild and does not affect the inference efficiency of a single agent.

\subsection{Analysis of Reasoning Distillation}
\label{sec-exp-case}

\textbf{Comparison with On-Policy Self-Distillation.}
We compare with on-policy self-distillation (SD)~\citep{zhao2026self}, where the model improves itself using its own generated trajectories.
Our PAD method outperforms SD (+0.99\% vs. +1.62\% on GSM8K).
The results highlight a fundamental distinction in supervision signals.
Specifically, SD relies on self-generated trajectories, which are constrained by the exploration capability of the student model; \method leverages teacher-induced MAS reasoning trajectories, providing richer and more structured supervision that captures diverse reasoning behaviors beyond the student’s own exploration.
Overall, these results suggest that MAS distillation offers a more informative and structured learning signal than purely on-policy self-distillation, leading to more effective reasoning transfer.

\textbf{Perplexity Analysis.}
We then conduct reasoning quality analysis by combining perplexity and LLM-based qualitative evaluation.
For each dataset, we randomly sample $100$ examples.
We measure the perplexity (\citep{bengio2003neural}; details in Appendix \ref{app:perplexity_cal}) of distilled models on held-out samples from GSM8K, focusing on reasoning tokens using Qwen3-32B and 0.6B as the teacher and student models, respectively.
Perplexity measures how well a model predicts the next token in a sequence; lower values indicate the model’s reasoning steps are more predictable and coherent, aligning with our goal of producing structured reasoning trajectories.
\Cref{tab:reasoning_quality} shows that all distilled models achieve substantially lower reasoning perplexity than the single ones, indicating that distillation improves the predictability of the student's reasoning.
This suggests that distillation not only helps produce more fluent outputs, but also encourages more structured and consistent reasoning trajectories.

\textbf{Evaluation of Reasoning Quality.}
For more quantitative evaluation, we employ InternLM-2.5-20b-chat \citep{cai2024internlm2} as an automatic evaluator to score outputs along four dimensions: step decomposition \citep{hwang2025assessing}, intermediate verification \citep{zheng2025beyond}, error localization \citep{mukherjee2025premise}, and overall reasoning coherence \citep{lee2025evaluating}.
Step decomposition evaluates whether the model explicitly breaks problems into logical substeps; intermediate verification measures self-checking at each step; error localization assesses the model’s ability to identify and correct mistakes; and overall reasoning coherence captures the consistency and logical flow of the solution.
As shown in \Cref{tab:reasoning_quality}, PAD achieves the highest scores in all metrics, indicating that it most effectively preserves explicit multi-step structure, self-checking behavior, and coherent reasoning flows.
In contrast, DA shows moderate improvements, particularly in intermediate verification and error localization, suggesting that it captures surface-level reasoning structure without fully inheriting reflective reasoning behaviors.
RSFT only outperforms the baseline, implying that direct reasoning-level supervision alone is insufficient.

\textbf{Case Study.}
We further presented detailed comparisons of reasoning states in Appendix \ref{app:case_study}.
This example demonstrates that the distilled single-agent model acquires the multi-agent reasoning patterns: its reasoning is more structured, logically coherent, and self-consistent, producing the correct answer without the repeated self-corrections seen in the baseline single-agent model.

\input{tables/tb-truth}

\subsection{Robustness and Generalization}
\label{sec-exp-robust}


\textbf{Robustness.}
We evaluate Qwen3-8B distilled from Qwen3-32B using the GSM8K dataset on TruthfulQA~\citep{lin2022truthfulqa}.
It measures the model’s ability to maintain factual accuracy and coherent reasoning.
As shown in \Cref{tab:truth_exp}, all distillation methods improve performance over the base model, indicating their robustness and resilience to catastrophic forgetting.
In particular, PAD achieves the highest scores across all metrics, indicating more robust reasoning and better retention of factual correctness.
These results suggest that MAS distillation not only enhances average accuracy but also increases robustness, allowing the student model to generalize more reliably to unseen or challenging tasks.

\begin{wrapfigure}{r}{0.5\textwidth}
    \centering
    \vspace{-.2in}
    \includegraphics[width=\linewidth]{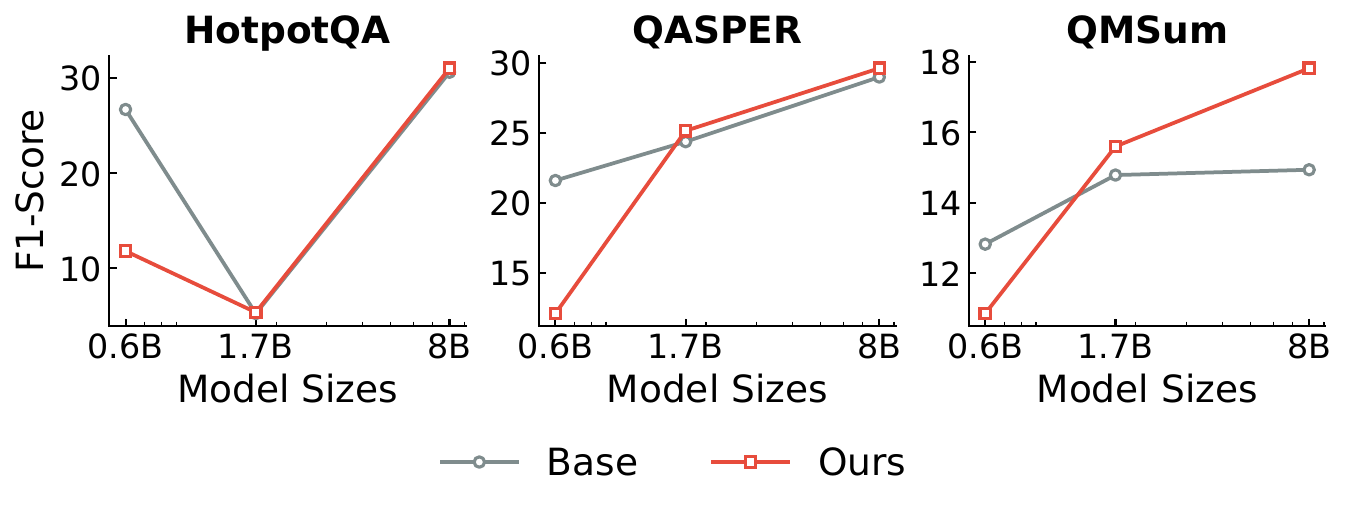}
    \vspace{-.3in}
    \caption{Performance on $3$ open-ended datasets.}
    \label{fig:generalizability_f1_scores}
    \vspace{-.2in}
\end{wrapfigure}

\textbf{Open-ended Generalization.}
We evaluate whether reasoning skills learned from mathematical problem-solving transfer to other complex reasoning tasks such as open-ended questions.
We train on a single source GSM8K and evaluate on three out-of-domain (OOD) datasets spanning diverse tasks: HotpotQA~\cite{yang2018hotpotqa} (multi-hop reasoning), QASPER~\cite{dasigi2021dataset} (long-context understanding), and QMSum~\cite{zhong2021qmsum} (summarization).
For comprehensive open-ended assessment, we leverage F1 scores and ROUGE-1/2/L~\cite{lin2004rouge} to measure lexical overlap between each prediction and the ground-truth, and BERTScore~\cite{zhang2019bertscore} for semantic similarity.
\Cref{fig:generalizability_f1_scores} shows that \method significantly enhances cross-dataset reasoning transfer, particularly for larger models, which consistently improve performance on diverse OOD tasks such as multi-hop QA, long-context understanding, and summarization.
This indicates that \method strengthens general reasoning capabilities rather than merely fitting dataset-specific pattern.
More details on other metrics are in Appendix~\ref{app:generalizability}.

\subsection{Extension to Multimodal LLMs, Heterogeneous Agents, and MAS Protocols}
\label{sec-exp-multi}

\begin{wrapfigure}{r}{0.5\columnwidth}
    \centering
    \vspace{-.25in}
    \includegraphics[width=\linewidth]{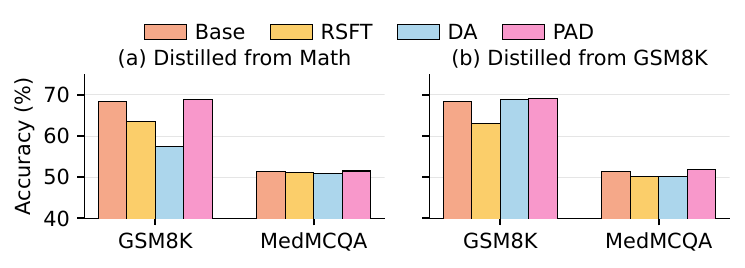}
    \vspace{-0.25in}
    \caption{Multimodal distillation results.}
    \label{fig:vlm_exp}
    \vspace{-0.2in}
\end{wrapfigure}
\textbf{Distillation to Multimodal LLMs (MLLMs).}
We perform distillation from Qwen2.5-VL-32B-Instruct into the smaller 3B version~\citep{bai2025qwen2}.
\Cref{fig:vlm_exp} shows that MAS distillation remains effective for MLLMs, despite being trained on text-only reasoning datasets.
Across both benchmarks, PAD consistently achieves the strongest or near-strongest performance, indicating that distilling process-level reasoning signals generalizes better than RSFT and DA.
Notably, the absolute gains are modest compared to text-only settings, which is expected since the model is not explicitly trained on MLLM reasoning.
This indicates that \method captures model-agnostic reasoning patterns that are transferable beyond the training modality.

\input{tables/generalization}

\textbf{Generalization across Multi-Agent Protocols.}
\method is designed to be algorithm-agnostic, as it distills reasoning dynamics from multi-agent interactions rather than relying on a specific coordination mechanism.
To validate this property, we extend beyond the standard debate setting and evaluate two additional MAS paradigms: Self-Consistency \citep{wang2022self} and AgentVerse \citep{chen2023agentverse}.
We distill from Qwen3-32B to Qwen3-8B using GSM8K as the training dataset, and evaluate on both GSM8K and MedMCQA.
As shown in \Cref{tab:mas_generalization}, \method consistently improves student performance across all MAS variants and distillation strategies, showing that our approach captures transferable reasoning patterns at the level of interaction dynamics, rather than overfitting to a particular multi-agent algorithm.

\textbf{Heterogeneous Settings.}
Finally, we study a more challenging \emph{heterogeneous} setting, where agents are instantiated from \textit{different} model families rather than a homogeneous backbone.
\Cref{tab:heterogeneous_results} shows that \method consistently improves performance over the single-agent baseline, demonstrating strong robustness to heterogeneous teacher compositions.
confirming that \method generalizes well across different multi-agent configurations.
More details are in Appendix~\ref{sec-app-hetero}.

\vspace{-.1in}
\section{Discussion}
\label{sec-disc}

Our experiments reveal several insights and avenues for future research.

\textit{(1) MAS distillation strategies:} Structured distillation, particularly PRM-guided methods, effectively transfers complex reasoning behaviors to smaller or multimodal models. Future work could explore adaptive distillation strategies that dynamically select reasoning trajectories based on task complexity, model capacity, or ensemble diversity. Additionally, methods that weigh intermediate self-corrections differently from final answers may further improve learning efficiency and reasoning fidelity.

\textit{(2) Process modeling and policy optimization:} Larger PRMs yield stronger gains even for small target models. Future research could investigate modular or hierarchical PRMs, which allow selective guidance for different reasoning components, and alternative policy optimization techniques that balance stability and scalability beyond PPO and GRPO.

\textit{(3) Scaling agent and data quality:} Increasing the number of debating agents benefits larger models but shows diminishing returns for smaller models. Similarly, training data volume alone is insufficient; high-quality reasoning trajectories are crucial. Future work could focus on adaptive ensemble scaling or selective trajectory sampling that optimizes the trade-off between diversity and learnability.


\textit{(4) Implication for foundation models:} Our study suggests that it is promising to leverage MAS to augment single LLMs, which will dramatically reduce cost with improved performance. Moreover, the distillation on small models implies that small language models can be significantly enhanced by MAS, highlighting promise for lightweight and cost-effective deployment of foundation models.

\textbf{Limitations.}
This work has the following limitations.
First, experiments are limited to a subset of reasoning benchmarks and multimodal models, evaluating additional tasks and modalities could help assess broader applicability.
Second, we focus on a specific set of distillation pipelines.
Exploring alternative or hybrid approaches may provide further insights.

\section{Conclusion}
We introduced \method, a framework for distilling multi-agent reasoning into a single agent.
Our experiments demonstrate that structured distillation, particularly PRM-guided methods, enables smaller models to approximate complex reasoning behaviors while maintaining efficiency and generalization across diverse tasks.
With the increasing attention to LLM efficiency and multi-agent systems, we hope our findings can provide insights for future research.

\section*{Acknowledgment}
This work was supported by Amazon Research Award, Spring 2025. Any opinions, findings, and conclusions or recommendations expressed in this material are those of the author(s) and do not reflect the views of Amazon. We also highlight the computing supported by Modal Academic Compute Grant and William \& Mary Research Computing for providing computational resources and/or technical support that have contributed to the results reported within this
paper. URL: https://www.wm.edu/it/rc.

\bibliographystyle{plain}
\bibliography{example_paper}


\appendix
\input{appendix}

\clearpage
\newpage
\input{checklist.tex}

\end{document}

%% file: tables/tb-truth.tex
\begin{table}[t!]
\centering
\caption{Evaluation of reasoning quality and robustness.}
\label{tab:combined_eval}
\vspace{-.1in}
\begin{subtable}{0.48\textwidth}
    \centering
    \caption{Quantitative evaluation of reasoning quality.}
    \label{tab:reasoning_quality}
    \vspace{-.1in}
    \setlength{\tabcolsep}{2.5pt}
    \resizebox{\linewidth}{!}{
    \begin{tabular}{lcccc}
    \toprule
    Metric & \textbf{Single} & \textbf{RSFT} & \textbf{DA} & \textbf{PAD} \\
    \midrule
    Avg NLL ($\downarrow$) & 0.6529 & 0.4092 & 0.4449 & 0.5876 \\
    Perplexity ($\downarrow$) & 1.9211 & 1.6388 & 1.5603 & 1.7996 \\
    \midrule
    Step Decomposition ($\uparrow$) & 2.75 & 3.13 & 3.38 & 3.23 \\
    Intermediate Verification ($\uparrow$) & 2.41 & 3.48 & 4.04 & 4.07 \\
    Error Localization ($\uparrow$) & 1.97 & 2.19 & 2.91 & 2.78 \\
    Reasoning Coherence ($\uparrow$) & 1.88 & 2.25 & 3.07 & 3.96 \\
    \bottomrule
    \end{tabular}
    }
\end{subtable}
\hfill
\begin{subtable}{0.48\textwidth}
    \centering
    \caption{Robustness evaluation on TruthfulQA.}
    \label{tab:truth_exp}
    \vspace{-.1in}
    \resizebox{\linewidth}{!}{
    \begin{tabular}{lcccc}
    \toprule
    Metric & \textbf{Single} & \textbf{PAD} & \textbf{DA} & \textbf{RSFT} \\
    \midrule
    BLEU ($\uparrow$)  & 0.6034 & 0.6634 & 0.6353 & 0.6059 \\
    ROUGE-1 ($\uparrow$) & 0.6144 & 0.6659 & 0.6389 & 0.6157 \\
    ROUGE-2 ($\uparrow$) & 0.5704 & 0.6414 & 0.5961 & 0.5777 \\
    ROUGE-L ($\uparrow$) & 0.6132 & 0.6573 & 0.6401 & 0.6157 \\
    \bottomrule
    \end{tabular}
    }
\end{subtable}
\vspace{-.2in}
\end{table}

%% file: tables/generalization.tex
\begin{table}[t]
\centering
\vspace{-5pt}
\caption{\method in different MAS protocols and heterogeneous settings.}
\vspace{-.1in}
\label{tab:mas_heterogeneous}
\small

\begin{subtable}{0.48\textwidth}
\centering
\caption{Performance in different MAS protocols.}
\vspace{-.1in}
\resizebox{\textwidth}{!}{
\begin{tabular}{lcccc}
\toprule
\textbf{MAS Method} & \textbf{Single} & \textbf{RSFT} & \textbf{DA} & \textbf{PAD} \\
\midrule
AgentVerse       
& 88.17 / 59.65 
& 88.12 / 59.70 
& 88.48 / 59.85 
& 88.53 / 59.78 \\
& 
& {\footnotesize \textcolor{cyan}{(-0.05 / +0.05)}} 
& {\footnotesize \textcolor{cyan}{(+0.31 / +0.20)}} 
& {\footnotesize \textcolor{cyan}{(+0.36 / +0.13)}} \\
Self-Consistency 
& 88.17 / 59.65 
& 88.20 / 59.58 
& 88.57 / 59.86 
& 88.58 / 59.81 \\
& 
& {\footnotesize \textcolor{cyan}{(+0.03 / -0.07)}} 
& {\footnotesize \textcolor{cyan}{(+0.40 / +0.21)}} 
& {\footnotesize \textcolor{cyan}{(+0.41 / +0.16)}} \\
\bottomrule
\end{tabular}
\vspace{-10pt}
}
\label{tab:mas_generalization}
\end{subtable}
\hfill
\begin{subtable}{0.42\textwidth}
\centering
\caption{Performance in heterogeneous settings.}
\vspace{-.1in}
\setlength{\tabcolsep}{3.5pt}
\resizebox{\textwidth}{!}{
\begin{tabular}{lcccc}
\toprule
\textbf{Student Model} & \textbf{Single} & \textbf{RSFT} & \textbf{DA} & \textbf{PAD} \\
\midrule
Qwen3-1.7B & 69.14 & 69.67 & 69.63 & 69.69 \\
Gemma-7B   & 51.10 & 52.95 & 53.77 & 53.21 \\
\bottomrule
\end{tabular}
}
\label{tab:heterogeneous_results}
\end{subtable}

\vspace{-20pt}
\end{table}

%% file: appendix.tex
\onecolumn

\appendix
\begin{center}
    {\large \textbf{Appendix}}
\end{center}



\begin{tcolorbox}[
    colback=gray!5,
    colframe=gray!40,
    boxrule=0.5pt,
    arc=2pt,
    left=10pt,
    right=10pt,
    top=8pt,
    bottom=8pt,
    title={\centering \textbf{Table of Contents}},
    fonttitle=\bfseries,
    coltitle=black,
    colbacktitle=gray!20
]
\newcommand{\appsection}[2]{\hyperref[#1]{\textbf{#2}}}
\newcommand{\appsubsection}[2]{\hspace{1.5em}\hyperref[#1]{#2}}

\begin{tabular}{@{}p{0.95\textwidth}@{}}
\appsection{app:data-generation}{A\quad Data Generation Methods and Pipeline} \\[2pt]
\appsubsection{app:dataset-overview}{A.1\quad Dataset Overview} \\
\appsubsection{app:multi-agent-data}{A.2\quad Multi-Agent Data Generation} \\
\appsubsection{app:correctness-filtering}{A.3\quad Correctness Filtering} \\
\appsubsection{app:data-selection}{A.4\quad Data Selection for Distillation with Data Augmentation} \\[6pt]

\appsection{app:methods}{B\quad Methods} \\[2pt]
\appsubsection{app:contrastive_loss}{B.1\quad Process Reward Model: Contrastive Loss Design} \\
\appsubsection{app:ppo-comparison}{B.2\quad Ablation: PPO Comparison} \\[6pt]

\appsection{app:sec-append-ablation}{C\quad Ablation Studies on PAD} \\[2pt]
\appsubsection{app:sec-append-ablation-prm}{C.1\quad PRM Modeling and Policy Separation} \\
\appsubsection{app:sec-append-ablation-ppo}{C.2\quad PPO vs.\ GRPO} \\[6pt]

\appsection{app:generalizability-main}{D\quad Generalization} \\[2pt]
\appsubsection{app:generalizability}{D.1\quad Generalization on Out-of-Domain Datasets} \\
\appsubsection{app:data-scaling}{D.2\quad Data Scaling} \\
\appsubsection{app:computation-cost}{D.3\quad Computation Cost} \\[6pt]

\appsection{app:case_study}{E\quad Case Study} \\[2pt]
\appsubsection{app:perplexity_cal}{E.1\quad Perplexity Calculation} \\[6pt]

\appsection{app:combine_methods}{F\quad Combined Distillation Strategies with Two-Step Training} \\[2pt]
\appsubsection{app:two_step}{F.1\quad Two-Step Training Protocol} \\[6pt]

\appsection{app:distillation-results}{G\quad Distillation Results} \\[2pt]
\appsubsection{app:sft_results}{G.1\quad Supervised Fine-tuning} \\
\appsubsection{app:all_models_results}{G.2\quad Reasoning Based Distillation Results} \\
\end{tabular}
\end{tcolorbox}

\section{Data Generation Methods and Pipeline}
\label{app:data-generation}

This section provides detailed descriptions of the data generation pipeline, dataset composition, and statistics used in our experiments.

\subsection{Dataset Overview}
\label{app:dataset-overview}

We construct the multi-agent distillation dataset using four core benchmarks, focusing on training the student models:
\begin{itemize}[leftmargin=2em]
\setlength\itemsep{0em}
    \item \textbf{Mathematical reasoning:} GSM8K and MATH, covering arithmatic, algebraic, and multi-step symbolic reasoning.
    \item \textbf{Augmented math reasoning:} MetaMathQA, providing additional multi-step problems and varied solution strategies.
    \item \textbf{Domain-specific knowledge:} MedMCQA, focusing on medical exam-style multiple-choice questions.
\end{itemize}

The remaining benchmarks, HotpotQA, QAPER, and QMSum, are reserved exclusively for zero-shot generalization evaluation and are not used during training or data augmentation.

For the training datasets, after multi-agent data generation, correctness filtering, and diversity-based selection, the final distillation dataset contains approximately 342k unique input questions and 2M reasoning trajectories.

\Cref{tab:dataset_statistics} reports detailed statistics for each training dataset, including the number of questions, retained debates, and augmented trajectories.

\subsection{Multi-Agent Data Generation}
\label{app:multi-agent-data}
To generate rich reasoning supervision, we adopt a multi-agent debate mechanism to produce diverse solution trajectories.
For each input problem $x$, we initialize a set of $n=5$ agents, plus a agent that summarize the final answer at the end.
\(
\mathcal{A} = \{a_1, a_2, \dots, a_n\}
\)
that share the same underlying teacher language model but operate with independent generation contexts.
Agents engage in a debate for up to $K=3$ rounds.
In the first round, each agent independently produces an initial reasoning trace and final answer.
In subsequent rounds, each agent observes the reasoning traces generated by other agents in the previous round and is encouraged to revise its own solution by identifying potential errors, alternative solution paths, or overlooked details.
Formally, in round $k$, agent $a_i$ generates a reasoning trajectory $\tau_{i,k}$ conditioned on the problem $x$ and the peer trajectories $\{\tau_{j,k-1}\}_{j \neq i}$.
The complete debate log for an input $x$ is denoted as
\[
\mathcal{L}_x = \{\tau_{i,k} \mid i \in [1,n], k \in [1,K]\}.
\]
This interaction protocol promotes reasoning behaviors such as self-correction, hypothesis revision, and cross-verification, yielding a diverse set of candidate solutions that go beyond single-pass generation.

\subsection{Correctness Filtering}
\label{app:correctness-filtering}

For each input problem \(x\), we collect responses generated by a set of agents \(\mathcal{A}\), where each agent produces a final answer along with its associated reasoning trajectory.
We first apply a \emph{correctness filtering} step to identify agents whose final answers are valid.

To ensure consistent and scalable evaluation across all tasks, we use Qwen2.5-72B-Instruct as an automatic verifier to assess whether a candidate final answer matches the ground-truth solution under task-specific evaluation rules (e.g., exact match or normalized equivalence for GSM8K-style problems).
This model is prompted to perform answer verification only and does not take the reasoning trajectory into account during this stage.

An agent \(a \in \mathcal{A}\) is included in the set of successful contributors \(\mathcal{A}_{\text{correct}}\) if its final answer is verified as correct by the verifier model.
This step ensures that all candidate responses used for distillation are answer-correct, independent of their reasoning styles.

Formally,
\begin{equation}
\mathcal{A}_{\text{correct}} =
\{ a \in \mathcal{A} \mid \text{Answer}(a, x) \text{ is verified as correct} \}.
\end{equation}

If \(|\mathcal{A}_{\text{correct}}| < 2\), the corresponding problem instance is excluded from the data augmentation process, as it does not provide sufficient diversity among correct solutions.

\input{tables/num_data}

\subsection{Data Selection for Distillation with Data Augmentation}
\label{app:data-selection}

\subsubsection{Divergent-Reasoning Selection via LLM-based Judgment}

Although all agents in \(\mathcal{A}_{\text{correct}}\) arrive at the same correct answer, their reasoning processes may differ substantially.
To capture this variation, we adopt a Correctness-First Diverse Extraction strategy that emphasizes reasoning divergence under answer agreement.

For each problem \(x\), we consider the set of reasoning traces produced by agents in \(\mathcal{A}_{\text{correct}}\).
We then employ Qwen2.5-72B-Instruct as an auxiliary judge to identify responses whose reasoning patterns are meaningfully distinct while preserving correctness.

The judge model is provided with the problem \(x\), the ground-truth answer, and multiple candidate reasoning traces from \(\mathcal{A}_{\text{correct}}\).
It is instructed to select responses that satisfy the following criteria:
\begin{enumerate}
    \item The final answer is correct.
    \item The reasoning process exhibits structural differences compared to other correct responses (e.g., different decomposition orders, intermediate representations, or solution paths).
\end{enumerate}

Notably, the judge model is not used to rank responses by quality or correctness. Instead, it operates solely to assess reasoning diversity conditioned on answer correctness, which reduces bias toward specific reasoning styles.

\subsubsection{Final Augmented Dataset Construction}

The selected responses are incorporated into the distillation dataset as augmented supervision signals.
By construction, the resulting dataset satisfies two properties:
\begin{itemize}
    \item \textbf{Answer consistency:} all augmented samples preserve the correct final answer.
    \item \textbf{Reasoning diversity:} multiple valid reasoning trajectories are retained for the same input.
\end{itemize}

This data selection procedure enables the student model to learn from a richer set of correct problem-solving behaviors without introducing incorrect or contradictory supervision.

\section{Methods}
\label{app:methods}

\subsection{Process Reward Model: Contrastive Loss Design}
\label{app:contrastive_loss}

Denoting $\sigma(\cdot)$ as the $\mathrm{sigmoid}$ function, we design the Process Reward Model (PRM) loss as a \textbf{contrastive loss} rather than standard binary cross-entropy. 
This choice encourages the model to assign higher rewards to reasoning steps that are more consistent with the multi-agent debate consensus, reflecting relative correctness rather than absolute labels.

Specifically, given a positive reasoning step $r_t^+$ and a set of negative steps $\{r_t^-\}$ sampled from other agents' reasoning trajectories, the PRM loss is defined as:

\begin{equation}
\mathcal{L}_{\text{PRM}}(\phi) = - \sum_t 
\log \mathrm{softmax}\Bigg(
\frac{\sigma(R_\phi(r_t^+)), \{\sigma(R_\phi(r^-))\}_{r^- \in \mathcal{N}_t}}{\tau}
\Bigg),
\end{equation}

where $\tau$ is a temperature hyperparameter controlling the sharpness of the contrastive distribution. 
This formulation encourages the PRM to score more consistent reasoning steps higher while down-weighting less consistent or contradictory steps.

\subsection{Ablation: PPO Comparison}
\label{app:ppo-comparison}
In addition to GRPO, we also experimented with standard Proximal Policy Optimization (PPO)~\citep{schulman2017proximal} to verify the benefits of group-wise relative updates.
The overall setup is identical to GRPO~\cite{shao2024deepseekmath} training, including the use of the Process Reward Model (PRM) to provide step-level supervision.
The student policy $\pi_\theta$ is fine-tuned using the conventional PPO clipped objective:

\begin{equation}
\mathcal{L}_{\text{PPO}}(\theta) = 
\mathbb{E}_{x \sim \mathcal{D}, o \sim \pi_{\text{old}}} \Big[ 
\min \big( \rho(\theta) \hat{A}, \text{clip}(\rho(\theta), 1-\epsilon, 1+\epsilon) \hat{A} \big) 
- \beta \mathbb{D}_{\text{KL}}(\pi_\theta \| \pi_{\text{ref}}) 
\Big],
\end{equation}

where $\rho(\theta) = \frac{\pi_\theta(o|x)}{\pi_{\text{old}}(o|x)}$ is the probability ratio and $\hat{A}$ is the advantage derived from PRM scores:

\begin{equation}
\hat{A} = \frac{R_{\phi}(o) - \mu_R}{\sigma_R},
\end{equation}

with $\mu_R$ and $\sigma_R$ denoting the mean and standard deviation of PRM scores across sampled outputs for a given input $x$.

Unlike GRPO, PPO treats each sample independently rather than comparing outputs within a group, which can lead to slower convergence and reduced reward consistency in multi-step reasoning tasks.
The comparison between GRPO and PPO in our ablation (Table~X) demonstrates that GRPO achieves more stable learning and higher final reasoning accuracy, validating the benefit of group-relative updates.

\section{Ablation Studies on PAD}
\label{app:sec-append-ablation}

\subsection{PRM Modeling and Policy Separation}
\label{app:sec-append-ablation-prm}

We study the role of PRM by explicitly separating the model used to learn process rewards from the model used to train the final policy.
In our setup, PRM is first trained independently, and its learned process-level guidance is then used to supervise the distillation of a target policy model.
All experiments distill from the same Qwen3-32B, while the PRM and the policy are instantiated with different parameter scales to analyze the effect of PRM capacity under this decoupled training scheme.
As shown in \Cref{tab:prm_ablation}, using a smaller PRM model consistently leads to limited improvement, even when paired with the same policy model.
In contrast, assigning a larger one yields gains across the GSM8K and MedMCQA test sets, even when the target policy remains lightweight.
These results indicate that effective distillation relies more critically on the modeling size of the PRM than on the scale of the policy itself, highlighting the importance of explicit process modeling and role separation.

\input{tables/tb-prm-llm}

\subsection{PPO vs. GRPO}
\label{app:sec-append-ablation-ppo}
To optimize the target policy during PRM-guided distillation, we compare Proximal Policy Optimization (PPO)~\citep{schulman2017proximal} (details in \Cref{app:ppo-comparison}) with GRPO.
PRMs in both settings are trained identically in a first stage and then fixed.
The two methods differ only in the second stage, where the target language model is fine-tuned using either PPO or GRPO under the same PRM-based reward signals.
As shown in \Cref{tab:ppo_ablation}, PPO achieves marginally higher accuracy in most settings.
This advantage is expected, as PPO employs a learned value function to provide a lower-variance, state-dependent baseline for policy updates, which can lead to more stable optimization under imperfect PRM rewards.
In contrast, GRPO removes the value function and relies on group-relative comparisons, resulting in slightly noisier updates but significantly reduced computational overhead.
Despite this difference, GRPO achieves performance comparable to PPO across all benchmarks, making it a scalable and effective alternative for large-scale PRM-guided distillation.

\input{tables/tb-ppo-grpo}

\section{Generalization}
\label{app:generalizability-main}

\subsection{Generalization on Out-of-Domain Datasets}
\label{app:generalizability}

\begin{figure}[h!]
    \centering
    \includegraphics[width=0.49\linewidth]{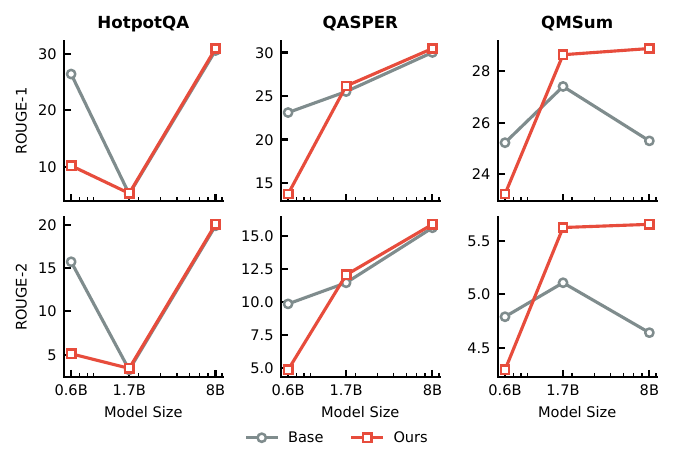}
    \includegraphics[width=0.49\linewidth]{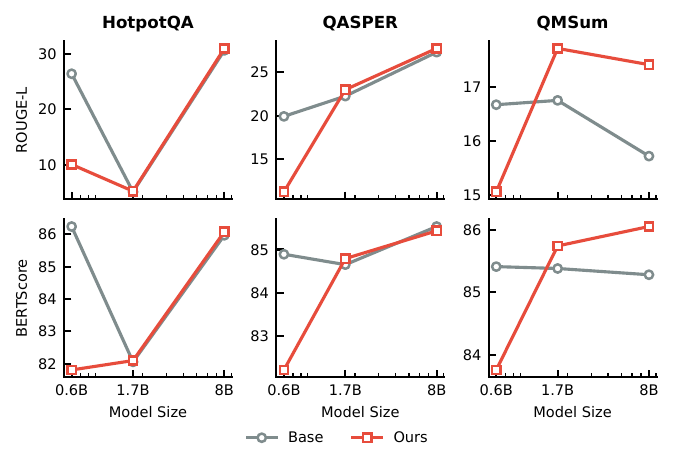}
    \caption{Scaling behavior of \method vs. base models. Performance comparison on three out-of-distribution (OOD) datasets across the Qwen model family demonstrate the impact of increasing model size ($0.6$B to $8$B) on generalizability. }
    \label{fig:generalizability_qwen_other_metrics}
\end{figure}

\begin{figure}[h!]
    \centering
    \includegraphics[width=0.5\linewidth]{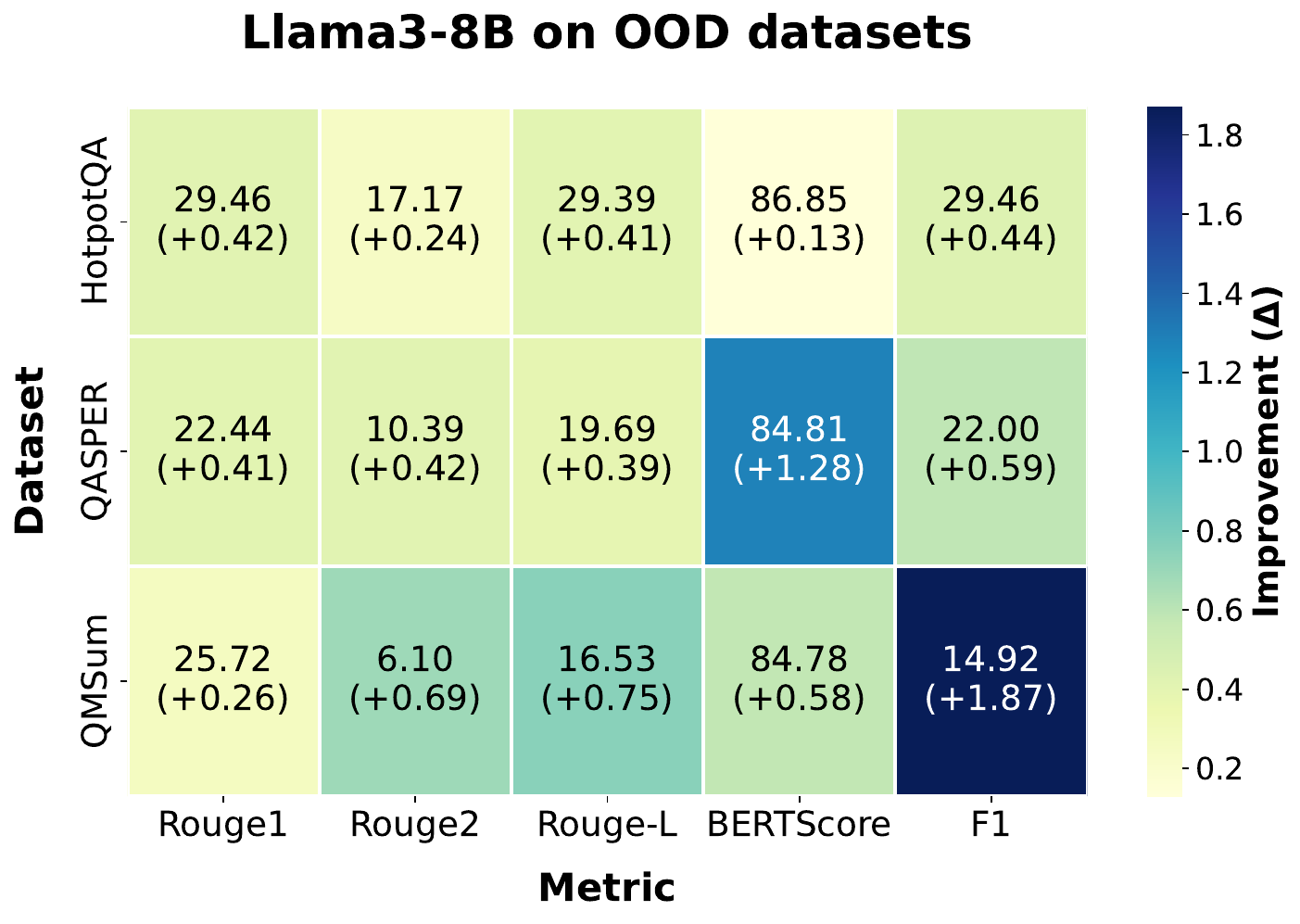}
    \caption{Performance of \method and the base models on out-of-domain datasets}
    \label{fig:generalizability_llama}
\end{figure}

To evaluate the robustness of \method, we conduct experiments across three out-of-domain (OOD) datasets spanning diverse domains: HotpotQA~\cite{yang2018hotpotqa} (multi-hop reasoning), QASPER~\cite{dasigi2021dataset} (long-context understanding), and QMSum~\cite{zhong2021qmsum} (summarization). 
The results in Figure~\ref{fig:generalizability_f1_scores} and Appendix Figure~\ref{fig:generalizability_qwen_other_metrics} \& \ref{fig:generalizability_llama} reveal a clear scaling law for reasoning transferability. 

The most substantial gains are observed in the larger 8B parameter class. Qwen3-8B and Llama3-8B demonstrate superior generalizability, effectively internalizing multi-agent thought processes to achieve consistent performance lifts across all OOD tasks. 
This `reasoning lift' is most pronounced in the complex QMSum task, where Qwen3-8B improves its F1-Score from $14.94$ to $17.82$ and its ROUGE-L Score from $15.72$ to $17.41$. Similarly, Llama3-8B sees its F1-Score increase from $13.05$ to $14.92$ on QMSum. These gains suggest that \method is particularly potent for complex tasks such as high-level synthesis and human-centered dialog understanding, where the model must perform multi-step reasoning and aggregate information from extensive documents multiple perspectives.

Even mid-sized models like Qwen3-1.7B begin to show this positive trajectory, with F1 scores rising from $24.38$ to $25.16$ on QASPER and from $14.79$ to $15.60$ on QMSum. 
This indicates that once a model surpasses a critical capacity threshold, \method serves as an enhancer for its inherent reasoning capabilities. In contrast, ultra-small models like Qwen3-0.6B have limited capacity to balance new reasoning patterns with its existing pre-trained knowledge base. Thus, \method's multi-agent knowledge is most effective for models with sufficient cognitive capacity.

\subsection{Data Scaling}
\label{app:data-scaling}
\begin{figure}[h!]
    \centering
    \includegraphics[width=0.35\textwidth]{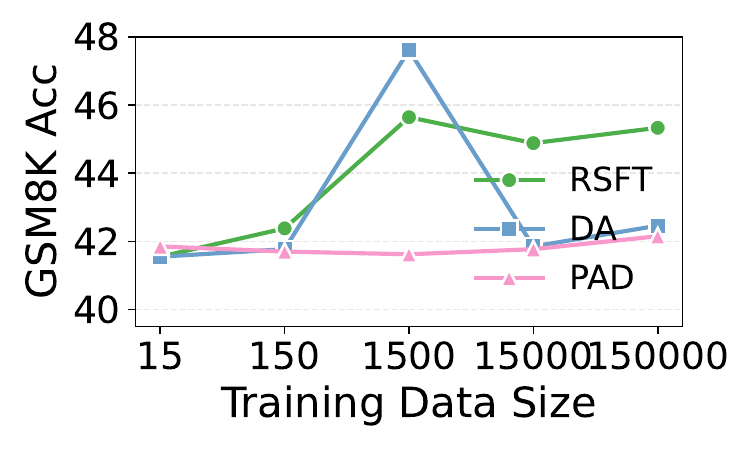}
    \caption{Data scaling behavior on MedMCQA for distillation from Qwen3-32B to Qwen3-0.6B, showing student model performance as a function of the amount of training data.}
    \label{fig:data_scale_add}
\end{figure}

We provide additional in \Cref{fig:data_scale_add} data-scaling results using the MedMCQA dataset, complementing the scaling analysis presented in \Cref{sec-exp-scaling} of the main paper.

\subsection{Computation Cost}
\label{app:computation-cost}
While AgentArk significantly reduces inference-time cost by eliminating multi-agent coordination, its training pipeline, particularly Process-Aware Distillation (PAD) introduces additional computational overhead.
We report concrete training costs for different distillation strategies to provide a realistic assessment of their practicality.

\paragraph{Experimental Setup.}
All experiments are conducted on NVIDIA H100 GPUs (80GB). 
Unless otherwise noted, we are reporting the student model has 8B parameters and is trained with a global batch size of 4 and identical sequence length across methods. 
All distillation approaches share the same multi-agent data generation cost, differences arise solely from the supervision signal and optimization procedure applied during student training.

\paragraph{Training Cost.}
\Cref{tab:cost_comparison} summarizes the additional training cost incurred by different distillation strategies.

\input{tables/computation_cost}

\paragraph{Inference-Time Cost.}
In contrast to the increased offline training cost, inference with AgentArk requires only a single autoregressive generation of the distilled student model. Compared to multi-agent debate, which typically involves multiple agent invocations and iterative coordination, AgentArk significantly reduces inference latency and GPU usage, making it suitable for real-time and resource-constrained deployment settings.

Overall, AgentArk trades increased offline computation for substantial reductions in inference-time cost and deployment complexity, consistent with common practices in large-scale model distillation and reinforcement learning.

\section{Case Study}
\label{app:case_study}








\begin{tcolorbox}[
      colback=gray!10,
      colframe=gray!50,
      width=\columnwidth, 
      height=5cm,
      title=Single Agent Reasoning
    ] \label{box:single}
Let's think step by step…  
Step 1: Green eggs = 5.  
Step 2: Blue eggs = 2 * 5 = 10.  
Step 3: Pink eggs = 10 - 1 = 9.  
Step 4: Yellow eggs = 1/3 * 9 = 3.  
Total = 5 + 10 + 9 + 3 = 27.  

Wait, but the answer is 24… let me check again…  \\
Recalculating yellow eggs: 9 / 3 = 3. Total still 27. Something is wrong…\\
Wait, but the answer is 24… let me check again…  \\
Recalculating yellow eggs: 9 / 3 = 3. Total still 27. Something is wrong…\\
Wait, but the answer is 24… let me check again…  \\
Recalculating yellow eggs: 9 / 3 = 3. Total still 27. Something is wrong…\\
...
\end{tcolorbox}

\begin{tcolorbox}[
      colback=gray!10,
      colframe=gray!50,
      width=\columnwidth, 
      height=5.2cm,
      title=Multi-Agent Reasoning
    ] \label{box:multi}
Agent 1:  
Step 1: Green eggs = 5  
Step 2: Blue eggs = 2 * 5 = 10  
Step 3: Pink eggs = 10 - 1 = 9  
Step 4: Yellow eggs = 9 / 3 = 3  
Total = 5 + 10 + 9 + 3 = 27 \\
Agent 2:  
Step 1: Green eggs = 5  
Step 2: Blue eggs = 10  
Step 3: Pink eggs = 9  
Step 4: Yellow eggs = 3  
Total = 27  
Verifies reasoning and intermediate calculations  \\
Agent 3:  
Step 1: Green = 5  
Step 2: Blue = 10  
Step 3: Pink = 9  
Step 4: Yellow = 3  
Total = 27 
Confirms each step independently…  \\
Agent 4:  
Step 1: Green = 5  
Step 2: Blue = 10  
Step 3: Pink = 9  
Step 4: Yellow = 3  
Total = 27
Checks reasoning consistency  \\
Agent 5:  
Step 1: Green = 5  
Step 2: Blue = 10  
Step 3: Pink = 9  
Step 4: Yellow = 3  
Total = 27 
All calculations verified  \\
Final Answer: 27  
\end{tcolorbox}

\begin{tcolorbox}[
      colback=gray!10,
      colframe=gray!50,
      width=\columnwidth, 
      height=5.8cm,
      title=Reasoning Comparison
    ] \label{box:ours}
\method : \\
The problem states that Cindy had 5 green eggs. \\
Step 1: She had twice as many blue eggs as green eggs, so she had 2 * 5 = 10 blue eggs. \\
Step 2: She had one fewer pink eggs than blue eggs, so she had 10 - 1 = 9 pink eggs. \\
Step 3: She had one-third as many yellow eggs as pink eggs, so she had 9 / 3 = 3 yellow eggs. \\
Step 4: The total number of eggs is 5 + 10 + 9 + 3 = 27 eggs. \\
Final Answer: 27\\
Reasoning: The problem states that Cindy had 5 green eggs. She had twice as many blue eggs as green eggs, so she had 2 * 5 = 10 blue eggs. She had one fewer pink eggs than blue eggs, so she had 10 - 1 = 9 pink eggs. She had one-third as many yellow eggs as pink eggs, so she had 9 / 3 = 3 yellow eggs. The total number of eggs is 5 + 10 + 9 + 3 = 27 eggs. The reasoning provided in the final answer is correct and matches the problem statement.
\end{tcolorbox}

\subsection{Perplexity Calculation}
\label{app:perplexity_cal}

\textbf{Definition.}
Given a reasoning sequence $x = (x_1, \dots, x_T)$, the perplexity (PPL) of a language model parameterized by $\theta$ is defined as:
\begin{equation}
\mathrm{PPL}(x) = \exp\left( - \frac{1}{T} \sum_{t=1}^{T} \log p_\theta(x_t \mid x_{<t}) \right),
\end{equation}
where $p_\theta(x_t \mid x_{<t})$ denotes the conditional probability assigned by the model to the next token $x_t$ given all previous tokens.

\textbf{Reasoning-Token Perplexity.}
Unlike standard perplexity evaluation over full responses, we restrict the computation to \emph{reasoning tokens} only.
Specifically, we identify reasoning spans corresponding to intermediate reasoning steps (e.g., chain-of-thought segments) and exclude prompt tokens, question descriptions, and final answer tokens.
Let $\mathcal{R} \subseteq \{1,\dots,T\}$ denote the index set of reasoning tokens.
The reasoning perplexity is then computed as:
\begin{equation}
\mathrm{PPL}_{\mathrm{reason}}(x) = 
\exp\left(
- \frac{1}{|\mathcal{R}|}
\sum_{t \in \mathcal{R}} 
\log p_\theta(x_t \mid x_{<t})
\right).
\end{equation}

\textbf{Evaluation Protocol.}
We evaluate perplexity on held-out GSM8K samples that are not used during distillation.
For each example, the gold reasoning trajectory is tokenized using the same tokenizer as the evaluated model.
The model is run in teacher-forcing mode to obtain token-level log-likelihoods.
Perplexity is computed per sample and then averaged across all evaluation samples.

\textbf{Model Setup.}
We compute perplexity for distilled student models and the single-agent baseline using the same teacher forcing procedure.
The Qwen3-32B model serves as the teacher, while Qwen3-0.6B is used as the student architecture.
All models are evaluated with identical prompts and decoding-free forward passes to ensure fair comparison.

\textbf{Interpretation.}
Lower reasoning perplexity indicates that the model assigns higher likelihood to coherent and structured reasoning steps, reflecting improved internal consistency and predictability of the reasoning process.
This aligns with our objective of distilling multi-agent reasoning dynamics into a single model.

\section{Combined Distillation Strategies with Two-Step Training}
\label{app:combine_methods}

In addition to evaluating individual distillation strategies, including reasoning-enhanced supervised fine-tuning (RSFT), reasoning data augmentation (DA), and process-aware distillation (PAD), we further investigate whether these methods can be effectively combined in a sequential manner.

\subsection{Two-Step Training Protocol}
\label{app:two_step}
We adopt a two-step training scheme that stacks DA on top of an existing distillation method. Specifically, we consider \textbf{RSFT+DA} and \textbf{PAD+DA}, where:
(i) in the first stage, the student model is trained using RSFT or PAD following the same setup as the main experiments; and
(ii) in the second stage, the resulting model is further fine-tuned with reasoning data augmentation to enhance reasoning diversity.

All experiments distill from a \textbf{Qwen3-32B} source model to a \textbf{Qwen3-1.7B} student model. In the second stage, we separately use reasoning data generated by the 32B model on \textbf{GSM8K} and \textbf{MATH}, denoted as \textit{32B\_gsm8k} and \textit{32B\_math}. The \textit{base} setting corresponds to using DA data aligned with the same task as the first-stage training.

Table~\ref{tab:mix_methods} summarizes the results of mixed distillation strategies on GSM8K and MedMCQA.

\input{tables/combine_methods}

Overall, stacking DA on top of RSFT or PAD leads to consistent but modest gains across benchmarks. 
While the improvements are incremental, the results suggest that our methods are mutually compatible, as reasoning data augmentation can be applied on top of different distillation strategies without degrading performance.

\section{Distillation Results}
\label{app:distillation-results}
\subsection{Supervised Fine-tuning}
\label{app:sft_results}
We evaluate the performance of distillation through standard SFT trained solely with ground-truth answer supervision, following common practice.
Specifically, models are fine-tuned using only input–output pairs from the ground-truth data, without any additional reasoning annotations, auxiliary losses, or trajectory-level supervision.
After training, we assess the performance both in-distribution performance and out-of-distribution generation under distribution shifts that require changes in reasoning.

The results is shown in \Cref{tab:sft_results}. 
A consistent pattern emerges across model families and scales. While SFT occasionally yields moderate gains on MedMCQA, it fails to produce consistent or reliable improvements on GSM8K, and in many cases leads to performance degradation.
Specifically, MedMCQA exhibits small but repeatable improvements when models are fine-tuned on medically oriented or structurally similar datasets (e.g., +6.8 for Gemma-7B, +3.4 for Qwen3-8B). 
This suggests that answer-only supervision can be beneficial when the target task shares surface-level structure or domain overlap with the training data, allowing models to exploit task-specific correlations.
In contrast, GSM8K performance remains largely unimproved or even deteriorates under SFT across nearly all settings. 
Even when trained directly on GSM8K, models frequently underperform their base counterparts.
This behavior indicates that SFT struggles to induce transferable reasoning strategies required for multi-step mathematical problem solving, and instead encourages overfitting to shallow input–output mappings that do not generalize beyond the supervised distribution.
Overall, these results highlight a key limitation of standard SFT: while answer-only supervision may yield localized gains on domain-specific benchmarks, it fails to support robust cross-task or reasoning-intensive generalization. 
This empirical evidence motivates the need for process-level supervision that exposes models to intermediate reasoning dynamics rather than only final outcomes.

\input{tables/sft_results}

\subsection{Reasoning Based Distillation Results}
\label{app:all_models_results}

Results for our three proposed methods are listed in \Cref{tab:qwen_results,tab:gemma_results,tab:llama_results,tab:qwen1_results,tab:qwen0_results}.

\input{tables/qwen_results}
\input{tables/llama_results}
\input{tables/gemma_results}
\input{tables/qwen1_results}
\input{tables/qwen0_results}

\subsection{\method in Heterogeneous Setting}
\label{sec-app-hetero}

We construct a MAS consisting of five heterogeneous agents: Qwen3-32B, Gemma-3-27B-IT, Devstral-Small-2-24B-Instruct-2512, QwQ-32B, and Phi-4-reasoning.
We evaluate two student models, Qwen3-1.7B and Gemma-7B, under this setting. The results are summarized in Table~\ref{tab:heterogeneous_results}. Across both students and all distillation strategies, \method consistently improves performance over the single-agent baseline, demonstrating strong robustness to heterogeneous teacher compositions.

In particular, for Qwen3-1.7B, RSFT, DA, and PAD improve performance by +0.53, +0.49, and +0.55 points respectively. For Gemma-7B, the gains are even more substantial, reaching +1.85, +2.67, and +2.11 points respectively.
These results indicate that heterogeneous teacher ensembles can provide complementary reasoning signals that further benefit distillation, especially for larger-capacity student models that can better absorb diverse supervision.

Overall, the performance trends are consistent with those observed in homogeneous MAS settings, confirming that \method generalizes well across different multi-agent configurations.
Importantly, the improvements remain stable across architectures and scales, suggesting that our approach effectively leverages diverse reasoning behaviors rather than relying on homogeneous consensus among agents.

\section{AI Assistants Usage}
\label{sec-app-llm}
AI assistants were used as auxiliary tools in preparing this manuscript, primarily for language refinement, clarity and organization. The experimental design and methodological choices were made by the authors.

\section{Broader Impact}
\label{sec-app-impact}

AgentArk offers meaningful societal benefits by enabling the advantages of multi agent reasoning through efficient distillation, rather than relying on expensive test time orchestration. This significantly reduces latency and deployment costs for reasoning intensive applications where traditional multi agent inference is impractical, such as on device or resource constrained environments. By lowering the computational barrier, AgentArk broadens access to advanced agentic reasoning capabilities, potentially democratizing their use across a wider range of users and applications.

However, these benefits are accompanied by potential risks. Distilled student models may inherit undesirable behaviors from teacher models, including biased, misleading, or logically unsound reasoning patterns that appear persuasive despite being incorrect. Such issues could lead to overreliance on flawed outputs, particularly in high stakes or decision support scenarios. To mitigate these concerns, we recommend incorporating rigorous correctness verification mechanisms for both process reward models (PRMs) and reinforcement learning fine tuned models, as well as systematically auditing distilled agents for hallucinations, bias, and harmful content prior to deployment.

Looking ahead, while AgentArk has the potential to extend to real world agentic settings, such as interactive tool use workflows and safety critical decision support tasks, careful consideration of reliability, transparency, and alignment will be essential to ensure responsible and trustworthy deployment.

%% file: tables/num_data.tex
\begin{table}[]
\centering
\caption{Dataset statistics for multi-agent distillation.
Q indicates the number of unique training questions, while T denotes the number of correct reasoning trajectories extracted per dataset.}
\label{tab:dataset_statistics}
\setlength{\tabcolsep}{2.5pt}
\resizebox{.7\textwidth}{!}{
\begin{tabular}{c|cccc}
\toprule
\textbf{Source Model} & \textbf{GSM8K(Q / T)} & \textbf{Math(Q / T)} & \textbf{MetaMathQA(Q / T)} & \textbf{Medmcqa(Q / T)} \\ \midrule
Qwen3-8B & 7473/44,838 & 500/3000 & 151k/906k & 183k/1.1M \\
Qwen3-32B & 7473/44,838 & 500/3000 & 151k/906k & 183k/1.1M \\
Gemma3-27B-It & 7473/44,838 & 500/3000 & 151k/906k & 183k/1.1M \\ \bottomrule
\end{tabular}
}
\vspace{-.1in}
\end{table}

%% file: tables/tb-prm-llm.tex
\begin{table}[htbp]
\centering
\caption{Ablation on PRM and LLM role separation.
All models are distilled from Qwen3-32B.}
\label{tab:prm_ablation}
\setlength{\tabcolsep}{2.5pt}
\resizebox{.45\textwidth}{!}{
\begin{tabular}{crr|ccccc}
\toprule
\multicolumn{1}{l}{\textbf{Test set}} & \multicolumn{1}{l}{\textbf{PRM}} & \multicolumn{1}{l|}{\textbf{LLM}} & \multicolumn{1}{l}{\textbf{GSM8K}} & \multicolumn{1}{l}{\textbf{MATH}} & \multicolumn{1}{l}{\textbf{MedMCQA}} & \multicolumn{1}{l}{\textbf{MMQA}} & \multicolumn{1}{l}{\textbf{Avg}} \\ \midrule
\multirow{4}{*}{GSM8K} & 0.6B & 0.6B & 42.84 & 42.68 & 42.46 & 42.53 & 42.63 \\
 & 8B & 0.6B & 42.84 & 42.53 & 42.91 & 42.23 & 42.63 \\
 & 0.6B & 8B & 88.48 & 88.55 & 88.40 & 88.63 & 88.52 \\
 & 8B & 8B & 88.63 & 88.70 & 88.48 & 88.56 & 88.59 \\ \midrule
\multirow{4}{*}{MedMCQA} & 0.6B & 0.6B & 32.39 & 32.58 & 32.36 & 32.54 & 32.47 \\
 & 8B & 0.6B & 32.49 & 32.56 & 32.35 & 32.20 & 32.40 \\
 & 0.6B & 8B & 59.50 & 59.74 & 59.74 & 59.65 & 59.66 \\
 & 8B & 8B & 59.77 & 59.81 & 59.86 & 59.77 & 59.80 \\ \bottomrule
\end{tabular}
}
\end{table}

%% file: tables/tb-ppo-grpo.tex
\begin{table}[htbp]
\centering
\caption{Ablation on PPO vs GRPO.}
\label{tab:ppo_ablation}
\setlength{\tabcolsep}{2.5pt}
\resizebox{.4\textwidth}{!}{
\begin{tabular}{lcccccc}
\toprule
 & \multicolumn{1}{l}{\textbf{Test set}} & \multicolumn{1}{l}{\textbf{GSM8K}} & \multicolumn{1}{l}{\textbf{MedMCQA}} & \multicolumn{1}{l}{\textbf{MMQA}} & \multicolumn{1}{l}{\textbf{Math}} & \multicolumn{1}{l}{\textbf{Avg}} \\ \midrule
\multirow{2}{*}{PPO} & GSM8K & 53.37 & 53.15 & 53.24 & 52.62 & 53.10 \\
 & MedMCQA & 49.56 & 50.08 & 49.77 & 49.41 & 49.71 \\ \midrule
\multirow{2}{*}{GRPO} & GSM8K & 52.71 & 52.48 & 52.62 & 52.64 & 52.61 \\
 & MedMCQA & 49.77 & 50.02 & 49.26 & 49.33 & 49.60 \\ \bottomrule
\end{tabular}
}
\vspace{-.25in}
\end{table}

%% file: tables/computation_cost.tex
\begin{table*}[htbp]
\centering
\caption{Training cost for different distillation strategies (8B student model).}
\label{tab:cost_comparison}
\resizebox{.8\textwidth}{!}{
\begin{tabular}{l l c c}
\toprule
\textbf{Method} & \textbf{Additional Training Components} & \textbf{GPUs} & \textbf{Time} \\
\midrule
RSFT 
& Supervised fine-tuning on single reasoning traces 
& 1 $\times$ H100 
& $\sim$6 hours \\

Reasoning DA 
& Supervised fine-tuning on augmented multi-trajectory reasoning data 
& 1 $\times$ H100 
& $\sim$8 hours \\

PAD (PRM + GRPO) 
& PRM training + GRPO-based policy optimization 
& 8 $\times$ H100 
& $\sim$20 hours \\

\quad PRM training 
& Step-level process reward modeling 
& 8 $\times$ H100 
& $\sim$8 hours \\

\quad GRPO
& Policy optimization with PRM reward 
& 8 $\times$ H100 
& $\sim$12 hours \\
\bottomrule
\end{tabular}
}
\end{table*}

%% file: tables/combine_methods.tex
\begin{table}[h]
\centering
\caption{Performance of combined distillation strategies with two-step training.}
\label{tab:mix_methods}
\setlength{\tabcolsep}{4pt}
\begin{tabular}{c|c|ccc}
\hline
\textbf{Method} & \diagbox{Test}{Train} & \textbf{Base} & \textbf{gsm8k} & \textbf{math} \\
\hline
\multirow{2}{*}{RSFT+DA}
& GSM8K     & 68.61 & 70.27 & 69.89 \\
& MedMCQA   & 42.67 & 43.92 & 44.30 \\
\hline
\multirow{2}{*}{PAD+DA}
& GSM8K     & 70.03 & 70.69 & 70.14 \\
& MedMCQA   & 42.72 & 43.68 & 43.49 \\
\hline
\end{tabular}
\end{table}

%% file: tables/sft_results.tex
\begin{table}[t!]
\centering
\caption{SFT performance}
\label{tab:sft_results}
\begin{tabular}{c|cccccc}
\hline
\diagbox{Test}{Train} & \textbf{} & \multicolumn{1}{l}{\textbf{base}} & \multicolumn{1}{l}{\textbf{gsm8k}} & \multicolumn{1}{l}{\textbf{medmcqa}} & \multicolumn{1}{l}{\textbf{metamathqa}} & \multicolumn{1}{l}{\textbf{math}} \\ \hline
gsm8k & \multicolumn{1}{c}{\multirow{2}{*}{llama3-8b}} & 75.28 & 63.61 & 73.31 & 65.78 & 75.36 \\
medmcqa & \multicolumn{1}{c}{} & 56.9 & 60.79 & 60.44 & 60.46 & 60.34 \\ \hline
gsm8k & \multirow{2}{*}{qwen3-8b} & 88.17 & 81.35 & 88.17 & 87.41 & 87.87 \\
medmcqa &  & 59.65 & 60.05 & 63.02 & 59.98 & 59.67 \\ \hline
gsm8k & \multirow{2}{*}{gemma-7b} & 51.1 & 56.56 & 48.52 & 36.39 & 50.04 \\
medmcqa &  & 49.29 & 48.98 & 57.06 & 48.15 & 48.43 \\ \hline
gsm8k & \multirow{2}{*}{qwen3-1.7b} & 69.14 & 59.59 & 70.66 & 66.34 & 67.32 \\
medmcqa &  & 42.34 & 43.72 & 49.53 & 43.68 & 42.46 \\ \hline
gsm8k & \multirow{2}{*}{qwen3-0.6b} & 41.93 & 37.91 & 44.35 & 21.38 & 31.31 \\
medmcqa &  & 32.2 & 32.9 & 38.9 & 34.09 & 32.27 \\ \hline
\end{tabular}
\end{table}

%% file: tables/qwen_results.tex
\begin{table}[]
\centering
\caption{Distillation results on Qwen3-8B.
Rows denote training datasets and columns denote test benchmarks (see diagonal header). Results compare RSFT, DA, and PAD across different source models. The original Qwen3-8B scores 88.17 on GSM8K and 59.65 on MedMCQA.}
\label{tab:qwen_results}
\setlength{\tabcolsep}{2.5pt}
\begin{tabular}{c|c|cccc}
\hline
\textbf{source model} & \diagbox{\textbf{test}}{\textbf{train}} & \textbf{gsm8k} & \textbf{medmcqa} & \textbf{metamathqa} & \textbf{math} \\ \hline
\multicolumn{1}{l|}{} & \multicolumn{1}{l|}{} & \multicolumn{4}{c}{RSFT} \\ \cline{3-6} 
\multirow{8}{*}{qwen3-8b} & gsm8k & 87.11 & 86.88 & 87.72 & 87.19 \\
 & medmcqa & 58.81 & 58.88 & 59.6 & 59.77 \\ \cline{3-6} 
 &  & \multicolumn{4}{c}{DA} \\ \cline{3-6} 
 & gsm8k & 86.43 & 86.35 & 88.1 & 86.41 \\
 & medmcqa & 58.33 & 58.83 & 57.49 & 58.35 \\ \cline{3-6} 
 &  & \multicolumn{4}{c}{PAD} \\ \cline{3-6} 
 & gsm8k & 88.42 & 88.36 & 88.48 & 88.41 \\
 & medmcqa & 59.71 & 59.81 & 59.71 & 59.73 \\ \hline
 &  & \multicolumn{4}{c}{RSFT} \\ \cline{3-6} 
\multirow{8}{*}{qwen3-32b} & gsm8k & 86.73 & 87.04 & 85.6 & 85.97 \\
 & medmcqa & 60.03 & 59.74 & 57.95 & 59.96 \\ \cline{3-6} 
 &  & \multicolumn{4}{c}{DA} \\ \cline{3-6} 
 & gsm8k & 87.11 & 87.49 & 89.57 & 86.2 \\
 & medmcqa & 59.48 & 58.69 & 58.33 & 59.86 \\ \cline{3-6} 
 &  & \multicolumn{4}{c}{PAD} \\ \cline{3-6} 
 & gsm8k & 89.05 & 89.02 & 89.15 & 88.7 \\
 & medmcqa & 60.04 & 63.12 & 61.53 & 61.21 \\ \hline
 &  & \multicolumn{4}{c}{RSFT} \\ \cline{3-6} 
\multirow{8}{*}{gemma3-27b-it} & gsm8k & 86.96 & 86.35 & 87.72 & 87.64 \\
 & medmcqa & 59.53 & 58.43 & 59.77 & 59.74 \\ \cline{3-6} 
 &  & \multicolumn{4}{c}{DA} \\ \cline{3-6} 
 & gsm8k & 87.79 & 86.28 & 86.2 & 86.05 \\
 & medmcqa & 59.43 & 58.5 & 57.18 & 59.72 \\ \cline{3-6} 
 &  & \multicolumn{4}{c}{PAD} \\ \cline{3-6} 
 & gsm8k & 88.48 & 88.48 & 88.48 & 88.4 \\
 & medmcqa & 59.96 & 59.84 & 59.86 & 59.79 \\ \hline
\end{tabular}
\end{table}

%% file: tables/llama_results.tex
\begin{table}[]
\centering
\caption{Distillation results on LLaMA3-8B-Instruct.
Rows denote training datasets and columns denote test benchmarks (see diagonal header). Results compare RSFT, DA, and PAD across different source models. The original LLaMA3-8B-Instruct scores 75.28 on GSM8K and 56.9 on MedMCQA.}
\label{tab:llama_results}
\setlength{\tabcolsep}{2.5pt}
\begin{tabular}{c|c|cccc}
\hline
\textbf{source model} & \diagbox{\textbf{test}}{\textbf{train}} & \textbf{gsm8k} & \textbf{medmcqa} & \textbf{metamathqa} & \textbf{math} \\ \hline
\multicolumn{1}{l|}{} & \multicolumn{1}{l|}{} & \multicolumn{4}{c}{RSFT} \\ \cline{3-6} 
\multirow{8}{*}{qwen3-8b} & gsm8k & 69.22 & 75.36 & 74.98 & 76.04 \\
 & medmcqa & 60.08 & 60.39 & 59.86 & 60.87 \\ \cline{3-6} 
 & \multicolumn{1}{l|}{} & \multicolumn{4}{c}{DA} \\ \cline{3-6} 
 & gsm8k & 70.89 & 75.06 & 73.24 & 75.82 \\
 & medmcqa & 56.44 & 58.36 & 58.59 & 56.83 \\ \cline{3-6} 
 & \multicolumn{1}{l|}{} & \multicolumn{4}{c}{PAD} \\ \cline{3-6} 
 & gsm8k & 76.42 & 75.28 & 75.55 & 75.72 \\
 & medmcqa & 57.11 & 57.16 & 57.02 & 57.1 \\ \hline
\multicolumn{1}{l|}{} & \multicolumn{1}{l|}{} & \multicolumn{4}{c}{RSFT} \\ \cline{3-6} 
\multirow{8}{*}{qwen3-32b} & gsm8k & 75.77 & 73.54 & 76.15 & 77.18 \\
 & medmcqa & 61.15 & 57.64 & 58.95 & 60.6 \\ \cline{3-6} 
 & \multicolumn{1}{l|}{} & \multicolumn{4}{c}{DA} \\ \cline{3-6} 
 & gsm8k & 77.8 & 74.39 & 77.53 & 75.88 \\
 & medmcqa & 56.47 & 58.12 & 58.83 & 56.77 \\ \cline{3-6} 
 & \multicolumn{1}{l|}{} & \multicolumn{4}{c}{PAD} \\ \cline{3-6} 
 & gsm8k & 76.88 & 75.97 & 75.89 & 76.95 \\
 & medmcqa & 60.82 & 60.75 & 60.79 & 60.96 \\ \hline
\multicolumn{1}{l|}{} & \multicolumn{1}{l|}{} & \multicolumn{4}{c}{RSFT} \\ \cline{3-6} 
\multirow{8}{*}{gemma3-27b-it} & gsm8k & 75.26 & 72.02 & 76.55 & 74.98 \\
 & medmcqa & 59.77 & 58.76 & 59.5 & 60.94 \\ \cline{3-6} 
 & \multicolumn{1}{l|}{} & \multicolumn{4}{c}{DA} \\ \cline{3-6} 
 & gsm8k & 77.89 & 72.33 & 76.37 & 75.59 \\
 & medmcqa & 56.44 & 55.3 & 58.93 & 56.92 \\ \cline{3-6} 
 & \multicolumn{1}{l|}{} & \multicolumn{4}{c}{PAD} \\ \cline{3-6} 
 & gsm8k & 76.35 & 75.97 & 77.02 & 76.27 \\
 & medmcqa & 60.82 & 60.94 & 60.73 & 60.55 \\ \hline
\end{tabular}
\end{table}

%% file: tables/gemma_results.tex
\begin{table}[]
\centering
\caption{Distillation results on Gemma-7B.
Rows denote training datasets and columns denote test benchmarks (see diagonal header). Results compare RSFT, DA, and PAD across different source models. The original Gemma-7B scores 51.1 on GSM8K and 49.29 on MedMCQA.}
\label{tab:gemma_results}
\setlength{\tabcolsep}{2.5pt}
\begin{tabular}{c|c|cccc}
\hline
\textbf{source model} & \diagbox{\textbf{test}}{\textbf{train}} & \textbf{gsm8k} & \textbf{medmcqa} & \textbf{metamathqa} & \textbf{math} \\ \hline
\multicolumn{1}{l|}{} & \multicolumn{1}{l|}{} & \multicolumn{4}{c}{RSFT} \\ \cline{3-6} 
\multirow{8}{*}{qwen3-8b} & gsm8k & 64.52 & 60.35 & 75.28 & 62.09 \\
 & medmcqa & 47.17 & 51.37 & 45.04 & 48.51 \\ \cline{3-6} 
 &  & \multicolumn{4}{c}{DA} \\ \cline{3-6} 
 & gsm8k & 65.73 & 62.02 & 74.75 & 62.62 \\
 & medmcqa & 46.38 & 50.97 & 42.84 & 48.84 \\ \cline{3-6} 
 &  & \multicolumn{4}{c}{PAD} \\ \cline{3-6} 
 & gsm8k & 52.74 & 51.69 & 52.34 & 52.55 \\
 & medmcqa & 49.32 & 49.66 & 49.37 & 49.42 \\ \hline
\multicolumn{1}{l|}{} & \multicolumn{1}{l|}{} & \multicolumn{4}{c}{RSFT} \\ \cline{3-6} 
\multirow{8}{*}{qwen3-32b} & gsm8k & 63.23 & 54.66 & 67.25 & 57.62 \\
 & medmcqa & 49.01 & 49.52 & 49.76 & 48.74 \\ \cline{3-6} 
 &  & \multicolumn{4}{c}{DA} \\ \cline{3-6} 
 & gsm8k & 66.26 & 58.23 & 73.78 & 63.46 \\
 & medmcqa & 48.31 & 48.06 & 48.74 & 49.32 \\ \cline{3-6} 
 &  & \multicolumn{4}{c}{PAD} \\ \cline{3-6} 
 & gsm8k & 63.07 & 63.37 & 63.6 & 62.46 \\
 & medmcqa & 49.7 & 50.77 & 49.61 & 50.77 \\ \hline
 &  & \multicolumn{4}{c}{RSFT} \\ \cline{3-6} 
\multirow{8}{*}{gemma3-27b-it} & gsm8k & 67.17 & 60.65 & 73.46 & 63.61 \\
 & medmcqa & 48.31 & 42.03 & 45.66 & 48.46 \\ \cline{3-6} 
 &  & \multicolumn{4}{c}{DA} \\ \cline{3-6} 
 & gsm8k & 68.16 & 60.27 & 71.34 & 59.59 \\
 & medmcqa & 47.36 & 38.35 & 42.34 & 48.7 \\ \cline{3-6} 
 &  & \multicolumn{4}{c}{PAD} \\ \cline{3-6} 
 & gsm8k & 53.37 & 53.15 & 53.24 & 52.62 \\
 & medmcqa & 50.62 & 50.08 & 50.77 & 49.41 \\ \hline
\end{tabular}
\end{table}

%% file: tables/qwen1_results.tex
\begin{table}[]
\centering
\caption{Distillation results on Qwen3-1.7B.
Rows denote training datasets and columns denote test benchmarks (see diagonal header). Results compare RSFT, DA, and PAD across different source models. The original Qwen3-1.7B scores 69.14 on GSM8K and 42.34 on MedMCQA.}
\label{tab:qwen1_results}
\setlength{\tabcolsep}{2.5pt}
\begin{tabular}{c|c|cccc}
\hline
\textbf{source model} & \diagbox{\textbf{test}}{\textbf{train}} & \textbf{gsm8k} & \textbf{medmcqa} & \textbf{metamathqa} & \textbf{math} \\ \hline
\multicolumn{1}{l|}{} & \multicolumn{1}{l|}{} & \multicolumn{4}{c}{RSFT} \\ \cline{3-6} 
\multirow{8}{*}{qwen3-8b} & gsm8k & 67.1 & 67.55 & 69.67 & 69.98 \\
 & medmcqa & 43.03 & 42.72 & 43.22 & 42.27 \\ \cline{3-6} 
 &  & \multicolumn{4}{c}{DA} \\ \cline{3-6} 
 & gsm8k & 68.99 & 69.07 & 69.29 & 69.23 \\
 & medmcqa & 48.49 & 45.18 & 42.65 & 42.41 \\ \cline{3-6} 
 &  & \multicolumn{4}{c}{PAD} \\ \cline{3-6} 
 & gsm8k & 69.82 & 69.41 & 69.78 & 69.76 \\
 & medmcqa & 42.36 & 42.42 & 42.41 & 42.44 \\ \hline
 &  & \multicolumn{4}{c}{RSFT} \\ \cline{3-6} 
\multirow{8}{*}{qwen3-32b} & gsm8k & 69.7 & 65.43 & 69.85 & 68.61 \\
 & medmcqa & 42.77 & 43.74 & 44.32 & 42.67 \\ \cline{3-6} 
 &  & \multicolumn{4}{c}{DA} \\ \cline{3-6} 
 & gsm8k & 70.78 & 68.84 & 73.89 & 70.03 \\
 & medmcqa & 43.68 & 42.46 & 43.03 & 42.72 \\ \cline{3-6} 
 &  & \multicolumn{4}{c}{PAD} \\ \cline{3-6} 
 & gsm8k & 70.36 & 70.05 & 70.28 & 69.83 \\
 & medmcqa & 44.41 & 44.46 & 42.43 & 42.46 \\ \hline
 &  & \multicolumn{4}{c}{RSFT} \\ \cline{3-6} 
\multirow{8}{*}{gemma3-27b-it} & gsm8k & 61.56 & 60.8 & 69.07 & 69.14 \\
 & medmcqa & 42.65 & 39.42 & 41.84 & 42.31 \\ \cline{3-6} 
 &  & \multicolumn{4}{c}{DA} \\ \cline{3-6} 
 & gsm8k & 71.11 & 65.06 & 69.94 & 64.52 \\
 & medmcqa & 43.99 & 40.69 & 42.58 & 42.34 \\ \cline{3-6} 
 &  & \multicolumn{4}{c}{PAD} \\ \cline{3-6} 
 & gsm8k & 69.67 & 70.2 & 69.98 & 69.9 \\
 & medmcqa & 42.36 & 42.53 & 42.46 & 42.55 \\ \hline
\end{tabular}
\end{table}

%% file: tables/qwen0_results.tex
\begin{table}[]
\centering
\caption{Distillation results on Qwen3-0.6B.
Rows denote training datasets and columns denote test benchmarks (see diagonal header). Results compare RSFT, DA, and PAD across different source models. The original Qwen3-0.6B scores 41.93 on GSM8K and 32.2 on MedMCQA.}
\label{tab:qwen0_results}
\setlength{\tabcolsep}{2.5pt}
\begin{tabular}{c|c|cccc}
\hline
\textbf{source model} & \diagbox{\textbf{test}}{\textbf{train}} & \textbf{gsm8k} & \textbf{medmcqa} & \textbf{metamathqa} & \textbf{math} \\ \hline
\multicolumn{1}{l|}{} & \multicolumn{1}{l|}{} & \multicolumn{4}{c}{RSFT} \\ \cline{3-6} 
\multirow{8}{*}{qwen3-8b} & gsm8k & 44.05 & 41.89 & 45.03 & 41.39 \\
 & medmcqa & 31.08 & 32.9 & 28.19 & 31.7 \\ \cline{3-6} 
 &  & \multicolumn{4}{c}{DA} \\ \cline{3-6} 
 & gsm8k & 43.67 & 42.23 & 51.78 & 41.17 \\
 & medmcqa & 28.76 & 30.29 & 30.27 & 30.72 \\ \cline{3-6} 
 &  & \multicolumn{4}{c}{PAD} \\ \cline{3-6} 
 & gsm8k & 42.57 & 42.25 & 42.51 & 42.53 \\
 & medmcqa & 32.43 & 32.33 & 32.38 & 32.27 \\ \hline
\multicolumn{1}{l|}{} & \multicolumn{1}{l|}{} & \multicolumn{4}{c}{RSFT} \\ \cline{3-6} 
\multirow{8}{*}{qwen3-32b} & gsm8k & 39.04 & 38.51 & 48.14 & 38.74 \\
 & medmcqa & 32.49 & 32.85 & 34.4 & 32.66 \\ \cline{3-6} 
 &  & \multicolumn{4}{c}{DA} \\ \cline{3-6} 
 & gsm8k & 48.67 & 40.94 & 53.37 & 41.55 \\
 & medmcqa & 34.62 & 32.06 & 34.38 & 34.54 \\ \cline{3-6} 
 &  & \multicolumn{4}{c}{PAD} \\ \cline{3-6} 
 & gsm8k & 42.84 & 42.46 & 42.53 & 42.68 \\
 & medmcqa & 33.09 & 34.36 & 34.54 & 32.58 \\ \hline
\multicolumn{1}{l|}{} & \multicolumn{1}{l|}{} & \multicolumn{4}{c}{RSFT} \\ \cline{3-6} 
\multirow{8}{*}{gemma3-27b-it} & gsm8k & 40.33 & 32.37 & 49.73 & 40.79 \\
 & medmcqa & 33.28 & 30.5 & 34.4 & 32.01 \\ \cline{3-6} 
 &  & \multicolumn{4}{c}{DA} \\ \cline{3-6} 
 & gsm8k & 41.02 & 42.46 & 49.2 & 40.94 \\
 & medmcqa & 32.27 & 32.94 & 35.43 & 33.23 \\ \cline{3-6} 
 &  & \multicolumn{4}{c}{PAD} \\ \cline{3-6} 
 & gsm8k & 44.61 & 43.52 & 43.72 & 42.17 \\
 & medmcqa & 33.01 & 34.51 & 33.27 & 32.92 \\ \hline
\end{tabular}
\end{table}

%% file: checklist.tex
\section*{NeurIPS Paper Checklist}

\begin{enumerate}

\item {\bf Claims}
    \item[] Question: Do the main claims made in the abstract and introduction accurately reflect the paper's contributions and scope?
    \item[] Answer: \answerYes{} 
    \item[] Justification: 
    \item[] Guidelines:
    \begin{itemize}
        \item The answer \answerNA{} means that the abstract and introduction do not include the claims made in the paper.
        \item The abstract and/or introduction should clearly state the claims made, including the contributions made in the paper and important assumptions and limitations. A \answerNo{} or \answerNA{} answer to this question will not be perceived well by the reviewers. 
        \item The claims made should match theoretical and experimental results, and reflect how much the results can be expected to generalize to other settings. 
        \item It is fine to include aspirational goals as motivation as long as it is clear that these goals are not attained by the paper. 
    \end{itemize}

\item {\bf Limitations}
    \item[] Question: Does the paper discuss the limitations of the work performed by the authors?
    \item[] Answer: \answerYes{} 
    \item[] Justification: See details in \Cref{sec-disc}
    \item[] Guidelines:
    \begin{itemize}
        \item The answer \answerNA{} means that the paper has no limitation while the answer \answerNo{} means that the paper has limitations, but those are not discussed in the paper. 
        \item The authors are encouraged to create a separate ``Limitations'' section in their paper.
        \item The paper should point out any strong assumptions and how robust the results are to violations of these assumptions (e.g., independence assumptions, noiseless settings, model well-specification, asymptotic approximations only holding locally). The authors should reflect on how these assumptions might be violated in practice and what the implications would be.
        \item The authors should reflect on the scope of the claims made, e.g., if the approach was only tested on a few datasets or with a few runs. In general, empirical results often depend on implicit assumptions, which should be articulated.
        \item The authors should reflect on the factors that influence the performance of the approach. For example, a facial recognition algorithm may perform poorly when image resolution is low or images are taken in low lighting. Or a speech-to-text system might not be used reliably to provide closed captions for online lectures because it fails to handle technical jargon.
        \item The authors should discuss the computational efficiency of the proposed algorithms and how they scale with dataset size.
        \item If applicable, the authors should discuss possible limitations of their approach to address problems of privacy and fairness.
        \item While the authors might fear that complete honesty about limitations might be used by reviewers as grounds for rejection, a worse outcome might be that reviewers discover limitations that aren't acknowledged in the paper. The authors should use their best judgment and recognize that individual actions in favor of transparency play an important role in developing norms that preserve the integrity of the community. Reviewers will be specifically instructed to not penalize honesty concerning limitations.
    \end{itemize}

\item {\bf Theory assumptions and proofs}
    \item[] Question: For each theoretical result, does the paper provide the full set of assumptions and a complete (and correct) proof?
    \item[] Answer: \answerNA{} 
    \item[] Justification: The paper does not present theoretical theorems or proof claims.
    \item[] Guidelines:
    \begin{itemize}
        \item The answer \answerNA{} means that the paper does not include theoretical results. 
        \item All the theorems, formulas, and proofs in the paper should be numbered and cross-referenced.
        \item All assumptions should be clearly stated or referenced in the statement of any theorems.
        \item The proofs can either appear in the main paper or the supplemental material, but if they appear in the supplemental material, the authors are encouraged to provide a short proof sketch to provide intuition. 
        \item Inversely, any informal proof provided in the core of the paper should be complemented by formal proofs provided in appendix or supplemental material.
        \item Theorems and Lemmas that the proof relies upon should be properly referenced. 
    \end{itemize}

    \item {\bf Experimental result reproducibility}
    \item[] Question: Does the paper fully disclose all the information needed to reproduce the main experimental results of the paper to the extent that it affects the main claims and/or conclusions of the paper (regardless of whether the code and data are provided or not)?
    \item[] Answer: \answerYes{} 
    \item[] Justification: We release the code
    \item[] Guidelines:
    \begin{itemize}
        \item The answer \answerNA{} means that the paper does not include experiments.
        \item If the paper includes experiments, a \answerNo{} answer to this question will not be perceived well by the reviewers: Making the paper reproducible is important, regardless of whether the code and data are provided or not.
        \item If the contribution is a dataset and\slash or model, the authors should describe the steps taken to make their results reproducible or verifiable. 
        \item Depending on the contribution, reproducibility can be accomplished in various ways. For example, if the contribution is a novel architecture, describing the architecture fully might suffice, or if the contribution is a specific model and empirical evaluation, it may be necessary to either make it possible for others to replicate the model with the same dataset, or provide access to the model. In general. releasing code and data is often one good way to accomplish this, but reproducibility can also be provided via detailed instructions for how to replicate the results, access to a hosted model (e.g., in the case of a large language model), releasing of a model checkpoint, or other means that are appropriate to the research performed.
        \item While NeurIPS does not require releasing code, the conference does require all submissions to provide some reasonable avenue for reproducibility, which may depend on the nature of the contribution. For example
        \begin{enumerate}
            \item If the contribution is primarily a new algorithm, the paper should make it clear how to reproduce that algorithm.
            \item If the contribution is primarily a new model architecture, the paper should describe the architecture clearly and fully.
            \item If the contribution is a new model (e.g., a large language model), then there should either be a way to access this model for reproducing the results or a way to reproduce the model (e.g., with an open-source dataset or instructions for how to construct the dataset).
            \item We recognize that reproducibility may be tricky in some cases, in which case authors are welcome to describe the particular way they provide for reproducibility. In the case of closed-source models, it may be that access to the model is limited in some way (e.g., to registered users), but it should be possible for other researchers to have some path to reproducing or verifying the results.
        \end{enumerate}
    \end{itemize}

\item {\bf Open access to data and code}
    \item[] Question: Does the paper provide open access to the data and code, with sufficient instructions to faithfully reproduce the main experimental results, as described in supplemental material?
    \item[] Answer: \answerYes{} 
    \item[] Justification: We release the code.
    \item[] Guidelines:
    \begin{itemize}
        \item The answer \answerNA{} means that paper does not include experiments requiring code.
        \item Please see the NeurIPS code and data submission guidelines (\url{https://neurips.cc/public/guides/CodeSubmissionPolicy}) for more details.
        \item While we encourage the release of code and data, we understand that this might not be possible, so \answerNo{} is an acceptable answer. Papers cannot be rejected simply for not including code, unless this is central to the contribution (e.g., for a new open-source benchmark).
        \item The instructions should contain the exact command and environment needed to run to reproduce the results. See the NeurIPS code and data submission guidelines (\url{https://neurips.cc/public/guides/CodeSubmissionPolicy}) for more details.
        \item The authors should provide instructions on data access and preparation, including how to access the raw data, preprocessed data, intermediate data, and generated data, etc.
        \item The authors should provide scripts to reproduce all experimental results for the new proposed method and baselines. If only a subset of experiments are reproducible, they should state which ones are omitted from the script and why.
        \item At submission time, to preserve anonymity, the authors should release anonymized versions (if applicable).
        \item Providing as much information as possible in supplemental material (appended to the paper) is recommended, but including URLs to data and code is permitted.
    \end{itemize}

\item {\bf Experimental setting/details}
    \item[] Question: Does the paper specify all the training and test details (e.g., data splits, hyperparameters, how they were chosen, type of optimizer) necessary to understand the results?
    \item[] Answer: \answerYes{} 
    \item[] Justification: We provide the information in \Cref{app:methods}
    \item[] Guidelines:
    \begin{itemize}
        \item The answer \answerNA{} means that the paper does not include experiments.
        \item The experimental setting should be presented in the core of the paper to a level of detail that is necessary to appreciate the results and make sense of them.
        \item The full details can be provided either with the code, in appendix, or as supplemental material.
    \end{itemize}

\item {\bf Experiment statistical significance}
    \item[] Question: Does the paper report error bars suitably and correctly defined or other appropriate information about the statistical significance of the experiments?
    \item[] Answer: \answerNo{} 
    \item[] Justification: Error bars are omitted due to the high computational cost of repeated experiments.
    \item[] Guidelines:
    \begin{itemize}
        \item The answer \answerNA{} means that the paper does not include experiments.
        \item The authors should answer \answerYes{} if the results are accompanied by error bars, confidence intervals, or statistical significance tests, at least for the experiments that support the main claims of the paper.
        \item The factors of variability that the error bars are capturing should be clearly stated (for example, train/test split, initialization, random drawing of some parameter, or overall run with given experimental conditions).
        \item The method for calculating the error bars should be explained (closed form formula, call to a library function, bootstrap, etc.)
        \item The assumptions made should be given (e.g., Normally distributed errors).
        \item It should be clear whether the error bar is the standard deviation or the standard error of the mean.
        \item It is OK to report 1-sigma error bars, but one should state it. The authors should preferably report a 2-sigma error bar than state that they have a 96\% CI, if the hypothesis of Normality of errors is not verified.
        \item For asymmetric distributions, the authors should be careful not to show in tables or figures symmetric error bars that would yield results that are out of range (e.g., negative error rates).
        \item If error bars are reported in tables or plots, the authors should explain in the text how they were calculated and reference the corresponding figures or tables in the text.
    \end{itemize}

\item {\bf Experiments compute resources}
    \item[] Question: For each experiment, does the paper provide sufficient information on the computer resources (type of compute workers, memory, time of execution) needed to reproduce the experiments?
    \item[] Answer: \answerYes{} 
    \item[] Justification: We provide the information in \Cref{app:computation-cost}
    \item[] Guidelines:
    \begin{itemize}
        \item The answer \answerNA{} means that the paper does not include experiments.
        \item The paper should indicate the type of compute workers CPU or GPU, internal cluster, or cloud provider, including relevant memory and storage.
        \item The paper should provide the amount of compute required for each of the individual experimental runs as well as estimate the total compute. 
        \item The paper should disclose whether the full research project required more compute than the experiments reported in the paper (e.g., preliminary or failed experiments that didn't make it into the paper). 
    \end{itemize}
    
\item {\bf Code of ethics}
    \item[] Question: Does the research conducted in the paper conform, in every respect, with the NeurIPS Code of Ethics \url{https://neurips.cc/public/EthicsGuidelines}?
    \item[] Answer: \answerYes{} 
    \item[] Justification: 
    \item[] Guidelines:
    \begin{itemize}
        \item The answer \answerNA{} means that the authors have not reviewed the NeurIPS Code of Ethics.
        \item If the authors answer \answerNo, they should explain the special circumstances that require a deviation from the Code of Ethics.
        \item The authors should make sure to preserve anonymity (e.g., if there is a special consideration due to laws or regulations in their jurisdiction).
    \end{itemize}

\item {\bf Broader impacts}
    \item[] Question: Does the paper discuss both potential positive societal impacts and negative societal impacts of the work performed?
    \item[] Answer: \answerYes{} 
    \item[] Justification: We offer in \Cref{sec-app-impact}
    \item[] Guidelines:
    \begin{itemize}
        \item The answer \answerNA{} means that there is no societal impact of the work performed.
        \item If the authors answer \answerNA{} or \answerNo, they should explain why their work has no societal impact or why the paper does not address societal impact.
        \item Examples of negative societal impacts include potential malicious or unintended uses (e.g., disinformation, generating fake profiles, surveillance), fairness considerations (e.g., deployment of technologies that could make decisions that unfairly impact specific groups), privacy considerations, and security considerations.
        \item The conference expects that many papers will be foundational research and not tied to particular applications, let alone deployments. However, if there is a direct path to any negative applications, the authors should point it out. For example, it is legitimate to point out that an improvement in the quality of generative models could be used to generate Deepfakes for disinformation. On the other hand, it is not needed to point out that a generic algorithm for optimizing neural networks could enable people to train models that generate Deepfakes faster.
        \item The authors should consider possible harms that could arise when the technology is being used as intended and functioning correctly, harms that could arise when the technology is being used as intended but gives incorrect results, and harms following from (intentional or unintentional) misuse of the technology.
        \item If there are negative societal impacts, the authors could also discuss possible mitigation strategies (e.g., gated release of models, providing defenses in addition to attacks, mechanisms for monitoring misuse, mechanisms to monitor how a system learns from feedback over time, improving the efficiency and accessibility of ML).
    \end{itemize}
    
\item {\bf Safeguards}
    \item[] Question: Does the paper describe safeguards that have been put in place for responsible release of data or models that have a high risk for misuse (e.g., pre-trained language models, image generators, or scraped datasets)?
    \item[] Answer: \answerNA{} 
    \item[] Justification: Paper poses no such risks.
    \item[] Guidelines:
    \begin{itemize}
        \item The answer \answerNA{} means that the paper poses no such risks.
        \item Released models that have a high risk for misuse or dual-use should be released with necessary safeguards to allow for controlled use of the model, for example by requiring that users adhere to usage guidelines or restrictions to access the model or implementing safety filters. 
        \item Datasets that have been scraped from the Internet could pose safety risks. The authors should describe how they avoided releasing unsafe images.
        \item We recognize that providing effective safeguards is challenging, and many papers do not require this, but we encourage authors to take this into account and make a best faith effort.
    \end{itemize}

\item {\bf Licenses for existing assets}
    \item[] Question: Are the creators or original owners of assets (e.g., code, data, models), used in the paper, properly credited and are the license and terms of use explicitly mentioned and properly respected?
    \item[] Answer: \answerYes{} 
    \item[] Justification: 
    \item[] Guidelines:
    \begin{itemize}
        \item The answer \answerNA{} means that the paper does not use existing assets.
        \item The authors should cite the original paper that produced the code package or dataset.
        \item The authors should state which version of the asset is used and, if possible, include a URL.
        \item The name of the license (e.g., CC-BY 4.0) should be included for each asset.
        \item For scraped data from a particular source (e.g., website), the copyright and terms of service of that source should be provided.
        \item If assets are released, the license, copyright information, and terms of use in the package should be provided. For popular datasets, \url{paperswithcode.com/datasets} has curated licenses for some datasets. Their licensing guide can help determine the license of a dataset.
        \item For existing datasets that are re-packaged, both the original license and the license of the derived asset (if it has changed) should be provided.
        \item If this information is not available online, the authors are encouraged to reach out to the asset's creators.
    \end{itemize}

\item {\bf New assets}
    \item[] Question: Are new assets introduced in the paper well documented and is the documentation provided alongside the assets?
    \item[] Answer: \answerNA{} 
    \item[] Justification:  The paper does not release new assets
    \item[] Guidelines:
    \begin{itemize}
        \item The answer \answerNA{} means that the paper does not release new assets.
        \item Researchers should communicate the details of the dataset\slash code\slash model as part of their submissions via structured templates. This includes details about training, license, limitations, etc. 
        \item The paper should discuss whether and how consent was obtained from people whose asset is used.
        \item At submission time, remember to anonymize your assets (if applicable). You can either create an anonymized URL or include an anonymized zip file.
    \end{itemize}

\item {\bf Crowdsourcing and research with human subjects}
    \item[] Question: For crowdsourcing experiments and research with human subjects, does the paper include the full text of instructions given to participants and screenshots, if applicable, as well as details about compensation (if any)? 
    \item[] Answer: \answerNA{} 
    \item[] Justification: The paper does not involve crowdsourcing nor research with human subjects.
    \item[] Guidelines:
    \begin{itemize}
        \item The answer \answerNA{} means that the paper does not involve crowdsourcing nor research with human subjects.
        \item Including this information in the supplemental material is fine, but if the main contribution of the paper involves human subjects, then as much detail as possible should be included in the main paper. 
        \item According to the NeurIPS Code of Ethics, workers involved in data collection, curation, or other labor should be paid at least the minimum wage in the country of the data collector. 
    \end{itemize}

\item {\bf Institutional review board (IRB) approvals or equivalent for research with human subjects}
    \item[] Question: Does the paper describe potential risks incurred by study participants, whether such risks were disclosed to the subjects, and whether Institutional Review Board (IRB) approvals (or an equivalent approval/review based on the requirements of your country or institution) were obtained?
    \item[] Answer: \answerNA{} 
    \item[] Justification: The paper does not involve crowdsourcing nor research with human subjects.
    \item[] Guidelines:
    \begin{itemize}
        \item The answer \answerNA{} means that the paper does not involve crowdsourcing nor research with human subjects.
        \item Depending on the country in which research is conducted, IRB approval (or equivalent) may be required for any human subjects research. If you obtained IRB approval, you should clearly state this in the paper. 
        \item We recognize that the procedures for this may vary significantly between institutions and locations, and we expect authors to adhere to the NeurIPS Code of Ethics and the guidelines for their institution. 
        \item For initial submissions, do not include any information that would break anonymity (if applicable), such as the institution conducting the review.
    \end{itemize}

\item {\bf Declaration of LLM usage}
    \item[] Question: Does the paper describe the usage of LLMs if it is an important, original, or non-standard component of the core methods in this research? Note that if the LLM is used only for writing, editing, or formatting purposes and does \emph{not} impact the core methodology, scientific rigor, or originality of the research, declaration is not required.
   \item[] Answer: \answerYes{} 
    \item[] Justification: Yes, in Appendix \Cref{sec-app-llm}
    \item[] Guidelines:
    \begin{itemize}
        \item The answer \answerNA{} means that the core method development in this research does not involve LLMs as any important, original, or non-standard components.
        \item Please refer to our LLM policy in the NeurIPS handbook for what should or should not be described.
    \end{itemize}

\end{enumerate}